\documentclass[12pt,twoside, a4paper1]{article}

\usepackage{natbib}
\usepackage{amsmath, amssymb, amsfonts}
\usepackage{mathtools}  
\usepackage{bm}         
\usepackage{physics}    
\usepackage{multirow}

\usepackage[margin=1in]{geometry}

\usepackage{graphicx}
\usepackage{float}       
\usepackage{subcaption}  

\usepackage{graphicx}
\usepackage{float}       
\usepackage{subcaption}  

\usepackage[colorlinks=true, linkcolor=blue, citecolor=blue, urlcolor=blue]{hyperref}
\usepackage[nameinlink]{cleveref}

\usepackage{enumitem}  
\usepackage[table, dvipsnames]{xcolor}

\usepackage{amsthm}
\usepackage{mdframed}
\usepackage{tikz}
\usetikzlibrary{calc}

\usepackage{amsthm}
\newtheorem{theorem}{Theorem}
\newtheorem{lemma}{Lemma}

\newtheorem{definition}{Definition}
\newtheorem{remark}{Remark}
\newtheorem{corollary}{Corollary}

\usepackage[colorlinks=true, linkcolor=blue, citecolor=blue, urlcolor=blue]{hyperref}
\usepackage[nameinlink]{cleveref}

\usepackage{soul} 

\usepackage{authblk}

\usepackage{cleveref} 

\usepackage{pifont}
\newcommand{\xmark}{\ding{55}}%
\newcommand{\zs}{z_\sigma}
\newcommand{\zb}{z_{\bar \sigma}}
\newcommand{\bsigma}{\bar \sigma}

\hypersetup{citecolor=MidnightBlue}

\newenvironment{bluebox}
{
\begin{mdframed}[hidealllines = true, backgroundcolor = Cerulean!10] \ \\[-0.35in] 
}
{
\end{mdframed}
}

\usepackage{setspace}

\title{\bf Generalized Guarantees for Variational Inference 
in the Presence of Even and Elliptical Symmetry}
\author[1]{Charles C. Margossian\thanks{\textit{Address for correspondence:} Charles Margossian, Earth Sciences Building, 2207 Main Mall Room 3182, Vancouver, British Columbia V6T 1Z4, Canada; email: charles.margossian@ubc.ca}}
\author[1]{Isaac E. Rankin}
\author[2]{Lawrence K. Saul}

\affil[1]{\small University of British Columbia, Department of Statistics, Vancouver, Canada}
\affil[2]{Flatiron Institute, Center for Computational Mathematics, New York, USA}

\date{ }

\begin{document}

\begin{singlespace}

\maketitle

\begin{abstract}
Variational inference (VI) approximates a target density $p$ by the best match $q$ in a family $\mathcal Q$ of tractable distributions.
The variational approximation is found by minimizing a divergence, most commonly the reverse Kullback-Leibler divergence $\text{KL}(q||p)$, although other divergences can be considered.
In practice, $p \notin \mathcal Q$ and so the optimal variational approximation is misspecified.
Despite this misspecification, it was recently shown that the minimizer of $\text{KL}(q||p)$ is guaranteed to recover the mean in the presence of even symmetry and likewise the correlation matrix in the presence of elliptical symmetry for a log-concave target $p$.
We extend this result in several directions.
First, we relax the assumption of log-concavity and instead merely require $p$ to be unimodal, allowing for heavy-tailed targets.
Second, we obtain guarantees for cases where $p$ exhibits symmetry along some \textit{but not all} of its coordinates.
These partial symmetries arise naturally in Bayesian hierarchical models, where the prior induces a challenging geometry but still possesses axes of symmetry.
Third, we extend these guarantees to $f$-divergences, a broad class which includes the forward and reverse-KL divergences and the $\alpha$-divergences.


\end{abstract}

\end{singlespace}



\section{Introduction}

Variational inference (VI) is a popular methodology for Bayesian inference and probabilistic machine learning 
\citep{Jordan:1999, Wainwright:2008, Blei:2017}.
The modus operandi of VI is to posit a family of tractable distributions~$\mathcal Q$ and find within this family the best approximation to a target distribution $p$ by minimizing a divergence.
By far the most common divergence to minimize is the reverse Kullback-Leibler (KL) divergence, although other options have been explored \citep[e.g.,][]{Wainwright:2008, Li:2016, Cai:2024}.
VI is typically presented as a scalable alternative to more classical algorithms, such as Markov chain Monte Carlo \citep[MCMC, see e.g.,][]{Robert:2004}.
While VI can quickly obtain an approximation within a constrained family~$\mathcal{Q}$,
it is often unclear how well this solution approximates $p$ \citep[e.g.,][]{Yao:2018, Giordano:2018, Talts:2018, Huggins:2020, Dhaka:2021}. For this reason, it is useful to understand the conditions under which VI returns
approximations that are \textit{provably} accurate. 
This paper contributes to a growing~body~of~work 
on this subject \citep[e.g.,][]{Wang:2018, Katsevich:2024, Margossian:2025robust,Marks:2026,Zellinger:2026}. 

In particular, we extend recent guarantees obtained when $\mathcal Q$ is a family of location-scale distributions.
Here it is known that, under certain regularity conditions, 
the minimizer of the reverse KL recovers both the mean and the correlation matrix of~$p$ whenever~$p$ is, respectively, even and elliptically symmetric, and furthermore $p$ is log-concave~\citep{Margossian:2025robust}.
The main results of our paper, contained in \Cref{thm:even-symmetry,thm:elliptical,thm:elliptical-sigma}, generalize these guarantees in several directions.
First, we relax the condition that $p$ is log-concave, requiring instead that $p$ is unimodal and allowing for cases where $p$ is heavy-tailed.
Second, we show that if $p$ exhibits symmetries in \textit{some but not all} of its coordinates, then VI still recovers the partial mean and correlations along these coordinates.
Of special note is that such partial symmetries arise in models with hierarchical priors.

Finally, our guarantees extend to $f$-divergences, a broad class of divergences that generalizes the reverse KL.
The $f$-divergences also include alternative objective functions for VI, such as as the forward KL \citep{Wainwright:2008, Naesseth:2020, Vehtari:2020} and the $\alpha$-divergences \citep{Li:2016, Dieng:2017, Daudel:2023},
which interpolate between the reverse and forward-KL as $\alpha$ moves from 0 to 1, and extrapolate when $\alpha \notin [0, 1]$. 
%
The extension of guarantees to $f$-divergences is primarily of theoretical interest, since alternatives to the reverse KL are notoriously difficult to optimize \citep{Geffner:2021}, although there 
are ongoing
efforts to develop adequate optimization strategies \citep[e.g.,][]{Daudel:2023b, Bertholom:2024}.

We note that our guarantees do not ensure 
exact recovery of the covariance matrix; rather, under elliptical symmetry, the covariance matrix is recovered \textit{up to a multiplicative positive scalar}, but this scalar may be arbitrarily small or large. 
In this sense, VI is not guaranteed to provide adequate uncertainty quantification.

\textbf{Related work.}
Our work most closely builds on guarantees for VI 
when minimizing the reverse KL-divergence for a log-concave target $p$, despite misspecifications in the family~$\mathcal Q$~\citep{Margossian:2025robust}.
In this paper we relax the assumption that $p$ is log-concave and instead require only that $p$ and $q$ are unimodal. Our guarantees with these milder assumptions extend the reach of many earlier results, although they do not provide a strict generalization; see \Cref{sec:related-guarantees} for a more detailed discussion.
Independently of our work, \citet{Zellinger:2026} studied conditions under which the minimizers of the forward KL and certain $\alpha$-divergences match the symmetry of $p$.
Their results are stronger in some respects and weaker in others. On one hand, they find
that VI can recover the mean of a \textit{mulitmodal} target $p$ when minimizing the forward KL or an $\alpha$-divergence with a more aggressive mass-covering behavior ($\alpha > 1$).
On the other hand, these guarantees, unlike our results, do not readily apply to the reverse KL, $\alpha$-divergences of order $\alpha \in (0, 1)$ or other $f$-divergences; see our discussion in \Cref{sec:related-guarantees}.

Concurrent to our work, \citet{Marks:2026} studied a more general class of symmetries and obtained guarantees \textit{under the assumption} that the $f$-divergence admits
a unique minimizer.
Their results are complementary to those in this paper,
where much of our technical contribution lies in identifying precisely those conditions under which an $f$-divergence admits a unique symmetry-matching minimizer or, in some cases, explicitly showing that all stationary points are symmetry-matching. 


More generally, our work contributes to a growing literature on VI. When minimizing the reverse-KL,
studies have demonstrated VI's ability to recover the mean, empirically \citep[e.g.,][]{Mackay:2003, Giordano:2018} and theoretically \citep[e.g.,][]{Katsevich:2024}, and others have obtained positive results when VI is used to maximize a marginal likelihood \citep{Jordan:1999, Li:2016} or construct frequentist estimators \citep{Wang:2018, Alquier:2020, Yang:2020, Zhang:2020}. 
On the other hand, many studies have proven negative results for VI, particularly when it is used to quantify uncertainty in~$p$ \citep{Mackay:2003, Turner:2011, Giordano:2018, Margossian:2023, Margossian:2025uncertainty}.
A complementary line of work examines post-hoc diagnostics to assess the quality of VI, for example using importance sampling \citep{Yao:2018, Vehtari:2024} or error bounds based on the Wasserstein distance~\citep{Huggins:2020, Biswas:2024}, and post-hoc corrections to improve variational approximations:
for example, \citet{Giordano:2018} examine a procedure to correct factorized approximations, and \citet{Pozza:2026} propose a skewing factor to correct symmetric approximations.

While less common, previous studies have also explored alternative objective functions for VI.
For example, when the forward KL-divergence is minimized, it is known that the optimal Gaussian approximation exactly recovers the target's mean and covariance \citep{Wainwright:2008}.
Studies have also shown that when different divergences are minimized, the approximations 
from VI yield different quantifications of uncertainty
and different estimators of the marginal likelihood \citep{Li:2016, Daudel:2023, Margossian:2025uncertainty}.
Our results provide a counterpoint of sorts:
we show that in the presence of certain symmetries, variational approximations from different divergences all recover the mean and the correlation matrix of~$p$, even when other properties of~$p$ are poorly estimated.


Finally, our study of partial symmetries relates to a large literature on the geometry of posteriors in hierarchical models and the interplay of this geometry with inference algorithms \citep[e.g.,][]{Neal:2001, Papaspiliopoulos:2007, Betancourt:2015}. 

\section{Preliminaries} \label{sec:preliminaries}

In this section, we provide formal definitions and identify the assumptions behind our theoretical analysis.

\subsection{Objective functions for VI}

\begin{table*}
\renewcommand{\arraystretch}{1.5}
\centering
  \begin{tabular}{l r c c c}
  \multirow{2}{*}{\bf Divergence} & 
  \multirow{2}{*}{\bf Notation} &
  \multirow{2}{*}{$f(u)$} & 
  {\bf Strictly} \\[-2ex] 
  & & & \hspace{1.5ex}{\bf convex}\hspace{1.5ex} & \\
  \rowcolor{Cerulean!10} (Reverse) Kullback-Leibler & $\text{KL}(q||p)$ & $- \log u$ & \checkmark \\ 
  $\alpha$-divergence, $\alpha \in [0, \infty) \backslash \{0, 1\}$ & $\text{D}_\alpha(p||q)$ & $\frac{u^\alpha - 1}{\alpha(\alpha -1)}$ &
  \checkmark \\ 
  \rowcolor{Cerulean!10} (Forward) Kullback-Leibler & $\text{KL}(p||q)$ & $u \log u$ & \checkmark \\ 
  Total variation distance (TVD) & $\text{TV}(p, q)$ & $\frac{1}{2}|u - 1|$ & \xmark \\ 
  \rowcolor{Cerulean!10} Smoothed TVD, $\beta > 0$  & $\text{TV}_{\beta}(p, q)$ & 
  $\frac{1}{2}\left(\sqrt{(u\!-\! 1)^2 + \beta^2} - \beta \right)$
  & \checkmark 
  \end{tabular}
  \caption{\textit{
  Example of $f$-divergences. 
  The $\alpha$-divergences include the squared Hellinger distance ($\alpha=1/2$) and the $\chi^2$-divergence ($\alpha=2$).
  We obtain our strongest guarantees when $f$ is strictly convex.
  }
  } \label{tab:divergences}
\end{table*}

VI minimizes a divergence between a target $p(z)$ and an approximation $q(z)$ over a family~$\mathcal Q$ of parameterized distributions.
We focus on the continuous case with $z \in \mathbb R^d$ and 
assume that both $p(z)$ and $q(z)$ admit a density with respect to a Lebesgue measure and have full support over $\mathbb R^d$.

There are many choices of divergences which can, at least in theory, be optimized for VI (\Cref{tab:divergences}).
The most common choice is the reverse KL divergence, which is often the easiest to minimize,
%
%
however several alternative divergences have been explored in attempts to capitalize on 
the attractive properties of their minimizers.
Certain algorithms approximately minimize the forward KL-divergence \citep{Naesseth:2020, Vehtari:2020}.
Others minimize $\alpha$-divergences \citep{Li:2016, Dieng:2017, Daudel:2023}; 
these are a broad class of divergences which interpolate between many well-known special cases, including 
the squared Hellinger distance when $\alpha=1/2$, the $\chi^2$-divergence when $\alpha=2$, and the reverse and forward KL-divergences in the limits $\alpha\rightarrow 0$ and $\alpha\rightarrow 1$, when the $\alpha$-divergence is defined as in~\citet{Cichocki:2010}.
%
Another useful measure of discrepancy between distributions is the total variation distance, used in the theoretical analysis of MCMC \citep[e.g.][]{Roberts:2004}.
Some of the 
objective functions we discuss here are not used for VI because they are too difficult to compute in high dimensions.
Even for these idealized objective functions, however, it remains of theoretical interest to understand when they yield an approximation $q$ that recovers statistical properties of~$p$.
We consider a broad class of divergences that includes all the divergences (and distances) described above.

\begin{bluebox}
\begin{definition}[\textbf{$f$-divergence \citep{Renyi:1961}}]
We refer to a divergence as an $f$-divergence if it can be written as
\begin{equation} \label{eq:f-divergence}
D_f(p||q) = \int f \left ( \frac{p(z)}{q(z)} \right) q(z) \text d z,
\end{equation}
where $f\!:\![0, \infty) \to \mathbb R$ is such that (i) $f$ is convex, (ii) $f(1) = 0$ and (iii) \mbox{$\text{lim}_{u \to 0^+} f(u) = f(0)$}.
\end{definition}
\end{bluebox}
Our theoretical analysis requires that the functions~$f$ is differentiable, an assumption that is tacitly made when practitioners use stochastic optimization in VI.

\subsection{Even, elliptical, and partial symmetries}

We focus on VI in settings where the target $p$ and the approximation $q$ exhibit certain symmetries. We begin by defining these symmetries for functions over $\mathbb{R}^d$ following \citet{Margossian:2025robust} and then we consider the special implications of these symmetries when they arise in densities.
For the latter, we leverage the established notion of elliptical distributions \citep[e.g.,][]{Fang:1990, Chmielewski:1981}.


\begin{bluebox}
\begin{definition}[\textbf{Even/odd symmetry}] 
We say a function $h: \mathbb R^d \to \mathbb R$ is even or odd symmetric about a point $\nu \in \mathbb R^d$ if, for all $\zeta \in \mathbb R^d$, it satisfies
\begin{align}
    \mbox{\textbf{even}}\quad\quad h(\nu+\zeta)\ &=\ +h(\nu - \zeta), \\
    \mbox{\textbf{odd}}\quad\quad h(\nu + \zeta)\ &=\ -h(\nu - \zeta).
\end{align}

\end{definition}
\end{bluebox}

\begin{bluebox}
\begin{definition}[\textbf{Elliptical symmetry}]
\label[definition]{def:elliptical}
We say that \mbox{$h\!:\!\mathbb R^d\! \to\! \mathbb R$} is elliptically symmetric about $\nu\!\in\!\mathbb R^d$ if there exists a positive-definite matrix $M\! \in\! \mathbb R^{d \times d}$, 
such that for any pair $z, z' \!\in\! \mathbb R^d$, 
we~have
\begin{equation}
h(z)=h(z')\quad \mbox{whenever}\quad (z\! -\! \nu)^T{M^{-1}}(z\!-\!\nu) = (z'\!-\!\nu)^T{M^{-1}}(z'\!-\!\nu).
\end{equation}
In the context where $h$ is the probability density of a distribution $p$, we call $M$ the scatter matrix of~$p$.
Note that the scatter matrix is only defined up to a positive scalar multiplier.
\end{definition}
\end{bluebox}

Since we assume that $p$ admits a density, we have that $p$ is an \textit{elliptical distribution} if it has an elliptically symmetric density.
We note however that the notion of elliptical distribution, and associated notions such as the scatter matrix, are more broadly defined using characteristic functions in order to accommodate cases where $p$ does not admit a density \citep[e.g.,][]{Fang:1990, Chmielewski:1981}.


Even and elliptical symmetries play a prominent role in probabilistic modeling for a number of reasons. One important reason is computational: when these symmetries arise in densities, they greatly simplify the calculation of certain statistics. 
For example, if $p(z)$ has a finite first moment and is even-symmetric about $\nu \in \mathbb R^d$, then the mean of $p$ is given by the point of symmetry, with $\mathbb E_p(z)=\nu$.
Similarly, if $p$ has finite second moments, then the scatter matrix $M$ from \Cref{def:elliptical} is proportional to the covariance matrix of $p$ and $p$ has correlation matrix,
\begin{equation} \label{eq:scatter-corr}
    \text{Corr}_p[z_i, z_j] = \frac{M_{ij}}{\sqrt{M_{ii} M_{jj}}}.
\end{equation}


The above properties reveal why symmetry-matching solutions are of great interest for VI. Namely, if a variational approximation $q$ matches the even symmetry of a target density $p$, then it correctly recovers the mean of $p$, and if it matches the ellipitical symmetry of $p$, then it correctly recovers the scatter matrix and hence the correlation matrix.
On the other hand, exact recovery of the scatter matrix does not imply exact recovery of the covariance matrix, since the scaling parameter may be arbitrarily small or large, and so symmetry does not ensure that VI correctly quantifies uncertainty.


\begin{table}
\renewcommand{\arraystretch}{1.5}
\begin{center}
    \begin{tabular}{l r r}
    \rowcolor{Cerulean!10}  {\bf Name} & {\bf Density kernel} & {\bf Radial function's tail} \\
    Gaussian & $\exp \left (- r^2 / 2 \right)$ & $\exp(-r^2 / 2)$ \\
    \rowcolor{Cerulean!10}  Laplace & $K_{d/2 -1}(r) / r^{d/2-1}$ & $\exp(-r)$ \\
    Student-t, $\delta > 0$ & $(1 + r^2/\delta)^{-(\delta + d)/2}$ & $ r^{-(\delta + d)}$ \\
    \end{tabular}
    \caption{\textit{Examples of standard elliptical distributions. 
    Above, $r = ||z-\nu||_{M{-1}}$, where $\nu$ is the point of even symmetry of $p$ and $M$ is its scatter matrix.
    In the kernel of the multivariate Laplace distribution, $K$ is a Bessel function of the second kind.
    $\delta$ is the degrees of freedom of the multivariate Student-t, and for $\delta=1$ we recover the multivariate Cauchy.
    The third column provides the asymptotic behavior of the radial function $g_p$ as $r$ becomes large.
    }} \label{tab:elliptical-distribution}
\end{center}
\end{table}

In a problem with elliptical symmetry, it can be simplifying to make a linear transformation to coordinates where the problem has spherical symmetry. 
Our next remark illustrates the form of this transformation for elliptical distributions and follows from the stochastic representation of elliptical distributions by~\citet{Fang:1990}.

\begin{bluebox}
\begin{remark}[\textbf{Radial function of elliptical density}] \label[remark]{remark:scalar_rep}
    Suppose $p$ is an elliptically symmetric density about $\mu$ with scatter matrix $M$, 
    and let $C$ be any matrix such that $M^{-1} = C^\top C$.
    Then there exists a scalar function $g_p: [0, \infty) \to [0, \infty)$ such that,
    \begin{equation}
        p(z) = g_p(||C(z-\mu)||)|C|,
   \end{equation}
    and we call this function the radial function of $p$. Moreover, the density of the transformed variable $\zeta=C(z\!-\!\mu)$ is spherically symmetric about the origin.
\end{remark}
\end{bluebox}

\vspace{-1ex}
In some cases, a density over $\mathbb{R}^d$ may only be symmetric along a particular subset of coordinates $\sigma\! \subset\! \{1,2,\ldots,d\}$.
We formalize this notion of partial symmetry below.

\vspace{1ex}
\begin{bluebox}
\begin{definition}[\bf Symmetry along $\sigma$]
Consider a distribution $p(\zs, \zb)$.
We say $p$ is even-symmetric along~$\sigma$ if for each $\zb$, $p(\zs|\zb)$ is even-symmetric about some point $\mu_\sigma(\zb)$.
We also say $p$ is elliptically symmetric along $\sigma$ if, for each $\zb$, $p(\zs|\zb)$ is elliptically symmetric with some scatter matrix $M_\sigma(\zb)$.
\end{definition}
\end{bluebox}

\begin{figure}[t]
    \centering
    \includegraphics[width=0.45\linewidth]{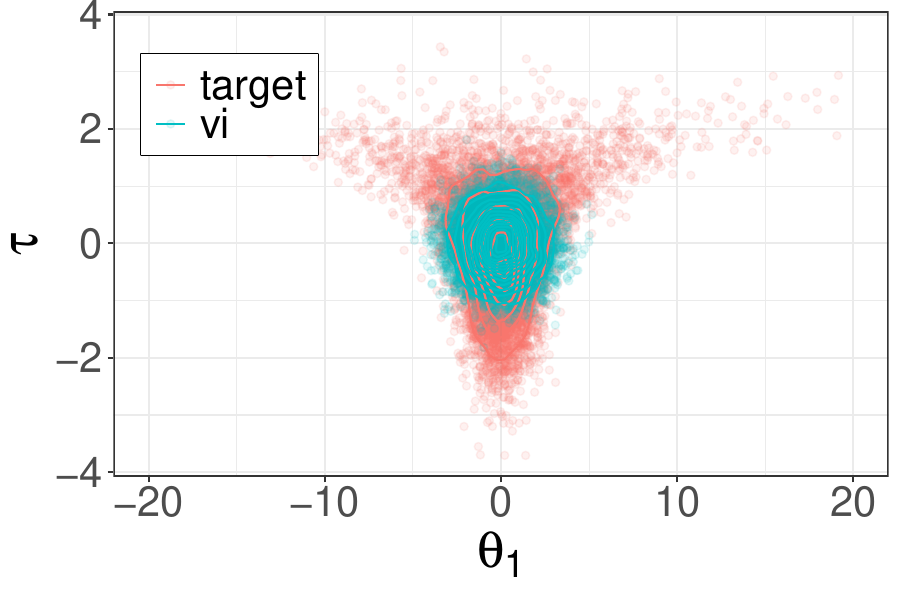}
    \includegraphics[width=0.45\linewidth]{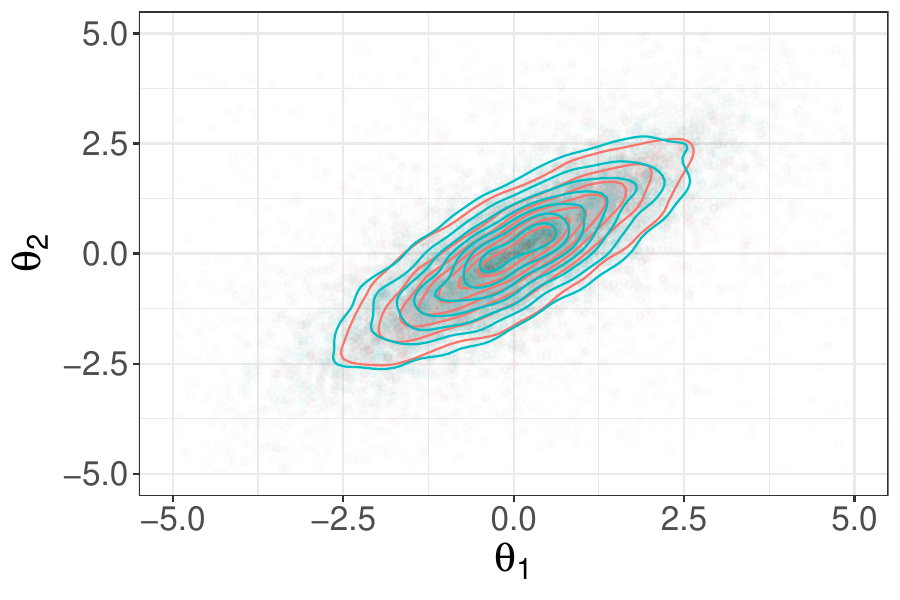}
    \caption{\textit{Gaussian VI approximation of an elliptical funnel,
    obtained by minimizing $\text{KL}(q||p)$.
    The funnel is asymmetric along $\tau$ but symmetric along $\theta$ and so VI provably recovers the mean and correlations of $\theta$.}}
    \label{fig:funnel}
\end{figure}

We provide an illustrative example of partial symmetry.
The elliptical funnel is the distribution over $\tau \in \mathbb R$ and $\theta \in \mathbb R^n$
generated by 
\begin{equation} \label{eq:funnel}
\tau\sim\mathcal{N}(0,1),\quad 
\theta \mid \tau \sim\mathcal{N}(0,e^{2\tau}M),
\end{equation}
where $M$ is a scatter matrix.
This is an extension of the well-known funnel \citep{Neal:2001}, in which $M$ is diagonal.
The geometry of the funnel is typical of hierarchical priors and prone to frustrate many inference algorithms. 
But the funnel also has partial symmetry:
the \textit{conditional} distribution $p(\theta|\tau)$ is elliptically symmetric with a point of even symmetry and a scatter matrix (proportional to $M$) that do not depend on the variable~$\tau$.
Later we will show that VI provably recovers the mean and the correlations along $\theta$ when $\mathcal Q$ is elliptically symmetric
(\Cref{fig:funnel}).

To match the symmetries of $p$, we need a family~$\mathcal{Q}$ whose distributions exhibit the same symmetries.
Natural choices for $\mathcal Q$ are the even-symmetric location family and the elliptical location-scale family.

\begin{bluebox}
\begin{definition}[\bf Even-symmetric location family]
    An even-symmetric location family $\mathcal Q$ is a one-parameter family of densities $q_\nu$ over $\mathbb R^d$, where $q_\nu$ is even-symmetric about $\nu \in \mathbb R^d$.
\end{definition} 
\end{bluebox}

\begin{bluebox}
\begin{definition}[\bf Elliptical location-scale family] \label{def:location-scale}
    An elliptical location-scale family $\mathcal{Q}$ is a two-parameter family of densities $q_{\nu, S}$ over $\mathbb R^d$,
    where $q_{\nu, S}$ is elliptically symmetric about $\nu \in \mathbb R^d$ with scatter matrix proportional to $S \in \mathbb R^{d\times d}$.
\end{definition}
\end{bluebox}

Elliptical location-scale families are prominent in probabilistic modeling.
Examples include the Gaussian, Laplace, Student-t, and Cauchy distributions.
They are also popular choices for VI, with the most standard choice being the family of Gaussians \citep[e.g.,][]{Ranganath:2014, Kingma:2014, Kucukelbir:2017, Cai:2024}.

\subsection{Regularity conditions on $p$ and $q$} \label{sec:regularity}

Our theoretical analysis is aided by imposing certain regularity conditions on $p$ and $q$.
First, for all of our results, we require that $p$ is differentiable, and more generally, 
that 
for each $q\!\in\!\mathcal{Q}$ we can 
interchange the order of differentiation and integration when differentiating $D_f(p||q)$.
The formal conditions for this assumption are provided by the dominated convergence theorem \citep[e.g.,][]{Billingsley:1995}.
This assumption is needed to minimize $D_f(p||q)$ via stochastic optimization.
%
Second, for some of our stronger results, we additionally
assume that the density $p(z)$ or $p(\zs|\zb)$ is strictly unimodal, as defined below.
\begin{bluebox}
\begin{definition}[\textbf{Unimodality of elliptical density}] \label{def:unimodal}
    Let $p$ be an elliptical distribution. 
    We say $p$ is unimodal if its radial function~$g_p$, as defined in \Cref{remark:scalar_rep}, is decreasing and strictly unimodal if $g_p$ is strictly decreasing.
\end{definition}
\end{bluebox}
%
In this paper, we show that when $p$ is\ strictly unimodal, any stationary point of $D_f(p||q)$ must be symmetry-matching.
We expect that this condition can be relaxed to non-strict unimodality but we do not provide a proof for this here and leave this extension to future work.


\section{Theoretical Guarantees} \label{sec:theory}

In this section, we provide broad conditions under which VI exactly recovers certain statistics of the target $p$ as a consequence of matching one or more of its underlying symmetries. 
Specifically, 
if $p$ admits a first moment, then by matching its even-symmetry VI exactly recovers the mean, 
and if $p$ admits a second moment, then by matching its elliptical symmetry VI exactly recovers the correlation matrix. 
In \Cref{sec:congruence}, we present a coordinate transformation that simplifies the minimization of $f$-divergences between elliptical distributions, and in \Cref{sec:even,sec:elliptical,sec:partial}, respectively, we prove the guarantees that follow from even, elliptical, and partial symmetries.

\subsection{Joint diagonalization by congruence}
\label{sec:congruence}

We derive a useful reformulation of the $f$-divergence when $p$ is elliptical and~$\mathcal Q$ is an elliptical location-scale family.
In this formulation, we pick a coordinate system that simultaneously diagonalizes the scatter matrices of $p$ and $q$ by congruence (\Cref{lemma:congruence}), and we characterize symmetry matching solutions (\Cref{lemma:stationary}) in this coordinate system.

\begin{bluebox}
\begin{lemma}[\textbf{Joint diagonalization by congruence}] \label[lemma]{lemma:congruence}
    Let $D_f$ be an $f$-divergence, and let $\mathcal Q$ be an elliptical location-scale family with location parameter $\nu$ and scale matrix $S$. Suppose $p$ is elliptically symmetric about $\mu$ with scatter matrix~$M$. Then for each $q\!\in\!\mathcal{Q}$, there exists a vector $\tau\!\in\!\mathbb{R}^d$ and a diagonal matrix $\Gamma\! \in\! \mathbb R^{d\times d}$ such that
    \begin{equation} \label{eq:Df_gl}
        D_f(p||q) = \int f \left ( \frac{g_p(||\Gamma (\zeta - \tau)||)}{g_q(||\zeta||)} |\Gamma| \right) g_q(||\zeta||)\, \text d \zeta.
    \end{equation}
\end{lemma}
\end{bluebox}

\begin{proof}
  Since $S$ and $M$ are positive-definite, they can be simultaneously diagonalized by congruence \citep{Horn:2012}. In particular, for any scale matrix $S$ of $q$, there exists a matrix $C$ and a positive-definite diagonal matrix $\Lambda$ such that 
   \begin{align}
      \label{eq:congS}
      CSC^\top &=\ I, \\
      \label{eq:congM}
      CMC^\top &=\ \Lambda. 
  \end{align}
  %
 From these congruence relations, we see that $C$ is a (full-rank) linear transformation that simultaneously transforms $q$ to a coordinate system where its scale matrix is the identity and $p$ to a coordinate system where its scatter matrix is $\Lambda$.
  Let $\zeta = C (z - \nu)$ denote these new coordinates, and observe per \cref{eq:congS} that
  $S^{-1} = C^\top C$.
  Then, per the definition of radial functions (\Cref{remark:scalar_rep}),
  \begin{equation} \label{eq:q-congruence}
      q(z) \text d z = g_q(||\zeta||) |C| |C|^{-1} \text d \zeta = g_q(||\zeta||) \text d \zeta.
  \end{equation}
  %
 


  Next, as useful notation, let $\Gamma\!=\!\Lambda^{-1/2}$, so that we rewrite the congruence relation in \cref{eq:congM} as  $M^{-1}\! =\! C^\top\Gamma^2 C$, and also let $\tau=C(\mu\!-\!\nu)$, so that we can write $z\!-\!\mu = C^{-1}(\zeta\!-\!\tau)$. Then again, per the definition of radial functions (\Cref{remark:scalar_rep}), we have
  \begin{equation} \label{eq:pq-congruence}
      \frac{p(z)}{q(z)} = 
      \frac{g_p(||\Gamma C (z\!-\!\mu)||)\,|\Gamma C|}{g_q(||C(z\!-\!\nu)||)\,|C|} = \frac{g_p(\|\Gamma(\zeta\! -\! \tau)\|)}{g_q(\|\zeta\|)} |\Gamma|.
  \end{equation}
  We recover \cref{eq:Df_gl} by substituting \cref{eq:q-congruence,eq:pq-congruence} into \cref{eq:f-divergence} for the $f$-divergence.
\end{proof}

\Cref{lemma:congruence} re-expresses the $f$-divergence in terms of the transformed variables $\tau$ and $\Gamma$.
This in turn lets us characterize symmetry-matching solutions, as done in the following lemma.

\begin{bluebox}
\begin{lemma}[\textbf{Symmetry-matching solutions after joint diagonalization}]
\label[lemma]{lemma:stationary}
Suppose the integral in \cref{eq:Df_gl} admits a stationary point at $(\tau, \Gamma) =(0, \gamma I)$ for some $\gamma > 0 $.
Then this stationary point corresponds to a stationary point $q$ of $D_f(p||q)$ that matches the even and elliptical symmetry of $p$.
\end{lemma}
\end{bluebox}
\begin{proof}

First we consider the even symmetry of $p$ and $q$. 
Since $C$ is full-rank and invertible, we have $\nu = \mu - C^{-1}\tau$.
Therefore, a minimizer of $D_f(p||q)$ with respect to $\tau$ produces a minimizer with respect to $\nu$.
Furthermore, we have that \textit{for any $S$}, $\tau = 0$ if and only if $\nu = \mu$, in which case $q$ matches the even symmetry of $p$.

Next we consider the elliptical symmetry of $p$ and $q$. We note that the congruence relationships in \cref{eq:congM,eq:congS} imply that $C$ and $\Gamma$ depend on $S$, even though we do not explicitly write these dependencies.
Observe now that if $\tau\!=\!0$, then the $f$-divergence as written 
in \cref{eq:Df_gl} 
depends on $S$ \textit{only} through~$\Gamma$.
In particular,
\begin{equation} \label{eq:S-dependence}
  \frac{\text d D_f}{\text d S} = \frac{\text d D_f}{\text d \Gamma} \frac{\text d \Gamma}{\text d S}. 
\end{equation}
Therefore, any stationary point with respect to $\Gamma$, satisfying $\text d D_f / \text d \Gamma\! =\! 0$, is also a stationary point of $D_f(p||q)$ with respect to $S$.
Moreover, we observed previously that  $S^{-1}\!=C^\top C$ and $M^{-1} \!= C^\top\Gamma^2 C$; thus if $\Gamma\!=\!\gamma I$, it follows at once that $S\!=\!\gamma^2 M$. Hence the scale matrix of $q$ is proportional to the scatter matrix of $p$, and $q$ matches the elliptical symmetry of~$p$.
\end{proof}

It now remains to show that under suitable conditions any minimizer of $D_f(p||q)$ indeed verifies $(\tau, \Gamma) = (0, \gamma I)$. This is the subject of the next sections.

\subsection{Guarantees from even symmetry}
\label{sec:even}

In this section, we generalize earlier guarantees for VI that, under the presence of even symmetry, a mean-matching solution must be a stationary point of $\text{KL}(q||p)$~\citep{Margossian:2025robust} to the broader family of $f$-divergences.
Furthermore, we provide conditions under which a symmetry-matching solution is a unique minimizer, requiring $p$ to be elliptically symmetric and unimodal, although not necessarily log-concave.

We begin with a lemma which shows that, under broad conditions, a solution that matches the point of even symmetry is a stationary point of $D_f(p||q)$.
\begin{bluebox}
\begin{lemma}[\textbf{Existence of an even symmetry-matching solution}] \label[lemma]{lemma:even-symmetry}
Let $\mathcal Q$ be an even-symmetric location family and let $D_f$ be an $f$-divergence.
If $p$ is even-symmetric about $\mu$, then a stationary point of $D_f(p||q_\nu)$ occurs at \mbox{$\nu\! =\! \mu$}.
\end{lemma}
\end{bluebox}

\begin{proof}
Write $q_\nu(z) = q_0(z\!-\!\nu)$ where by assumption $q_0$ is even-symmetric about the origin. Then with the change of variables $\zeta=z\!-\!\nu$, we can rewrite the $f$-divergence in \cref{eq:f-divergence} as
\begin{equation}
  D_f(p||q_\nu) = \int\! f\! \left( \frac{p(\zeta + \nu)}{q_0(\zeta)} \right ) q_0(\zeta) \text d \zeta.
\end{equation}
To determine where $D_f$ is stationary, we now differentiate with respect to the location parameter $\nu$ and carry the gradient through the integral,
%
%
\begin{equation}
  \nabla_\nu D_f(p||q_\nu)\,
     =\, \int\! f'\! \left( \frac{p(\zeta + \nu)}{q_0(\zeta)} \right ) \nabla_\nu p(\zeta + \nu)\, \text d \zeta\, 
     =\,  \int\! f'\! \left( \frac{p(\zeta + \nu)}{q_0(\zeta)} \right ) \nabla_\zeta p(\zeta + \nu)\, \text d \zeta,
\end{equation}
where in the last equality, we have used the symmetry of $\nu$ and $\zeta$ in the argument of $p$ to take the gradient with respect to $\zeta$ rather than $\nu$.
If we set $\nu\! =\! \mu$, then \mbox{$p(\nu\! +\! \zeta)$} is even-symmetric about the origin and $\nabla_\zeta p(\nu\! +\! \zeta)$ is odd symmetric.
All other terms in the integrand are even-symmetric; hence the integral vanishes, indicating that a stationary point of $D_f(p||q_\nu)$ occurs at $\nu\!=\!\mu$.
\end{proof}

In general, this stationary point is not a global minimizer; see for a counter-example a balanced mixture of two Gaussians where the symmetry-matching solution is a local \textit{maximizer} \citep{Margossian:2025robust}.
To obtain guarantees for a unique solution when minimizing the reverse KL, \citet{Margossian:2025robust} restrict $p$ to be log-concave.
However, this condition excludes cases where $p$ is heavy-tailed, for example, the Student-t whose density is not log-concave but merely quasi-concave.
Furthermore, the resulting proof does not generalize to $f$-divergences because, even if $p$ is log-concave, $f(p(\zeta + \nu)/q_0(\zeta))$ is in general not convex in $\nu$ (the reverse KL is an exception).

Here, we propose a set of conditions for the uniqueness of the stationary point, which broadly holds across $f$-divergences and applies to heavy-tailed distributions.
Specifically, we require $p$ to be a unimodal elliptical distribution and similarly, $\mathcal Q$ to be a family of elliptical location-scale distributions.
We also require $f$ to be strictly convex, a condition that is verified by all the $f$-divergences in \Cref{tab:divergences}, except the total variation distance.
However, in \Cref{tab:divergences} we also define the \textit{smoothed total variation distance}, 
%
%
which is a valid $f$-divergence for any $\beta >  0$, with $f$ strictly convex and can be made arbitrarily close (pointwise) to the total variation distance by taking $\beta \to 0$. 

\begin{bluebox}
    \begin{theorem}[\textbf{Recovery of even symmetry}] \label{thm:even-symmetry}
Let $D_f$ be an $f$-divergence, let $p$ be an elliptically symmetric distribution with point of even symmetry $\mu$, and let $\mathcal Q$ be an elliptical location-scale family with location parameter $\nu$.
In addition, suppose that $f$ is strictly convex, and that $p$ and every $q \in \mathcal Q$ are strictly unimodal.
Then $D_f(p||q)$ has a unique minimizer which occurs at \mbox{$\nu\! =\! \mu$}.
\end{theorem}
\end{bluebox}

\begin{proof}

From \Cref{lemma:even-symmetry}, we know that $\nu\!=\!\mu$ is a stationary point.
We will now show that if $\nu\!\neq\!\mu$, then we can find a non-zero directional derivative in a direction that moves $\nu$ closer to $\mu$ (as measured by some Mahalanobis distance based on the scatter matrices of $p$ and $q$).

We start with the formulation of the $f$-divergence, introduced in \Cref{lemma:congruence} and restated here for convenience,
\begin{equation*}
  D_f(p||q) = \int f \left ( \frac{g_p(||\Gamma (\zeta\!-\!\tau)||)}{g_q(||\zeta||)} |\Gamma| \right) g_q(||\zeta||)\, \text d \zeta,
\end{equation*}
where $\tau = C(\mu - \nu)$ and, as shown in \Cref{lemma:stationary}, a minimizer of $D_f(p||q)$ with respect to $\tau$ produces a minimizer with respect to $\nu$.
Throughout, we consider that $\tau\neq0$, as implied by $\nu \neq \mu$.
Taking the gradient with respect to $\tau$, we obtain
\begin{align}
    \nabla_\tau D_f(p||q) 
    & = \int f' \left ( \frac{g_p(||\Gamma (\zeta\!-\!\tau)||)}{g_q(||\zeta||)} |\Gamma| \right) |\Gamma| \frac{g_p'(||\Gamma(\zeta - \tau)||)}{||\Gamma (\zeta - \tau)||} \Gamma^2 (\tau - \zeta)  \text d \zeta
    \label{eq:gradient-u}
\end{align}
where 
we have used the handy result that $\nabla_\tau||\Gamma(\zeta-\tau)|| = \Gamma^2(\tau-\zeta) / ||\Gamma(\zeta-\tau)||$.
Let $v \in \mathbb R^d$ be a direction, which in time we will define.
To study the sign of the directional derivative, $\nabla_\tau D_f(p||q) \cdot v$, we split the state space into subspaces, based on whether the non-radial term in the integrand is positive or negative.
In particular, let $\Omega^+ = \left \{ \zeta: (\tau-\zeta)^T\Gamma^2v > 0 \right \}$
and similarly let $\Omega^-$ and $\Omega^0$ be the sets where the left side of the above inequality is respectively strictly negative and zero. 
A prescient choice for the direction $v$ is given by $v = \Gamma^{-2} \tau$.
Then,
\begin{equation} \label{eq:direction-gradient}
    \nabla_\tau D_f(p||q) \cdot v = \int f' \left ( \frac{g_p(||\Gamma (\zeta\!-\!\tau)||)}{g_q(||\zeta||)} |\Gamma| \right) |\Gamma| \frac{g_p'(||\Gamma(\zeta - \tau)||)}{||\Gamma (\zeta - \tau)||} (\tau - \zeta)^T \tau \  \text d \zeta,
\end{equation}
and $\Omega^0 = \{\zeta: (\tau - \zeta)^T\tau = 0 \}$ is now an affine hyperplane containing $\tau$, while $\Omega^+$ and $\Omega^-$ partition $\mathbb R^d$ (\Cref{fig:omega-partition}).

The plan for the remainder of the proof is as follows.
We pair each point $\zeta \in \Omega^+$ with a point $\zeta' \in \Omega^-$, obtained by taking the reflection of $\zeta$ about $\tau$. 
We then re-express the integral over $\mathbb R^d$ in \cref{eq:direction-gradient} as a \textit{folded} integral over $\Omega^+$ and show that the resulting integrand is strictly positive for any $\zeta \in \Omega^+$.
Therefore, the integral itself does not vanish, and so $\tau \neq 0$ cannot be a stationary point.
We refer the reader to \Cref{app:proof-even-symmetry} for the remaining details of the proof; there we
also highlight why $p$ is required to be unimodal.
\end{proof}

\begin{figure}
\begin{center}
\begin{tikzpicture}[scale=2]

\draw[->] (-0.5,0) -- (2,0) node[right] {$\zeta_1$};
\draw[->] (0,-0.5) -- (0,2) node[above] {$\zeta_2$};

\coordinate (O) at (0,0);

\coordinate (U) at (0.8,0.6);

\draw[->, thick] (O) -- (U) node[midway, above left] { };

\draw[dashed] ($(U) + (-1.2,1.6)$) -- ($(U) + (0.9,-1.2)$) node[right] {$\Omega_0$};

\node at (0.7, -0.5) {$\textcolor{MidnightBlue}{\Omega^+}$};
\node at (1.5, 0.5) {$\textcolor{Orange}{\Omega^-}$};

\coordinate (Z) at (0.7, 0.2);
\fill (Z) circle (1pt) node[right] {$\textcolor{MidnightBlue}{\zeta}$};

\coordinate (Zprime) at (0.9, 1);

\fill (Zprime) circle (1pt) node[left] {\textcolor{Orange}{$\zeta'$}};

\draw[dotted] (Z) -- (Zprime);

\fill (U) circle (1pt) node[above right] {$\tau$};
\end{tikzpicture}
\end{center}
\caption{
\textit{For each point $\textcolor{MidnightBlue}{\zeta \in \Omega^+}$, we obtain a reflection $\textcolor{Orange}{\zeta' \in \Omega^-}$.
Exploiting the even-symmetry of $p$ and the spherical symmetry of $q$ (after the right coordinate transformation), we show that the sum of the integrand $\nabla_\tau D_f(p||q) \cdot v$ evaluated at $\zeta$ and $\zeta'$ is always strictly positive, and so the directional derivative $\nabla_\tau D_f(p||q)\cdot v$ must be non-zero, unless $\tau=0$.
}
} \label{fig:omega-partition}
\end{figure}

\begin{figure}[t]
\centering
    \includegraphics[width=.27\linewidth]{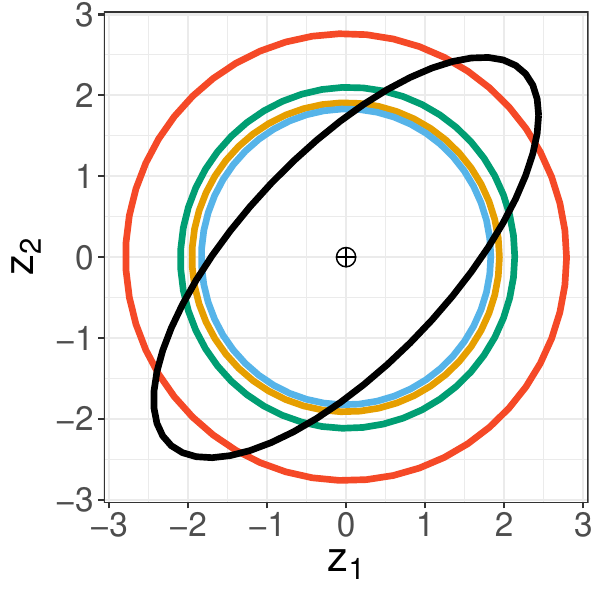} \ 
    \includegraphics[width=.4\linewidth]{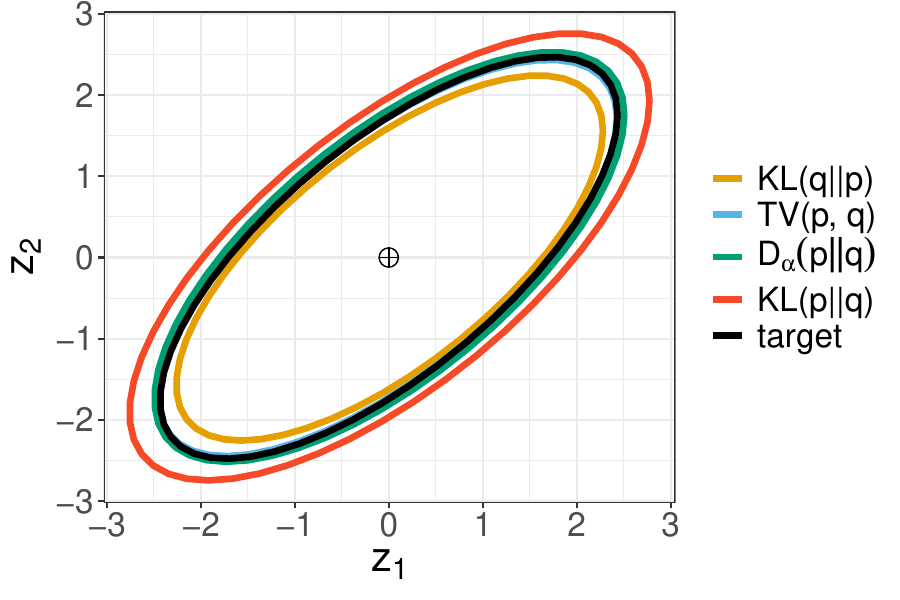}
    \caption{\textit{Variational approximation of a multivariate Student-t by a Gaussian. For each divergence in \Cref{tab:divergences}, the mean of the Student-t is recovered by a factorized Gaussian approximation (left), while its correlation matrix is recovered by a non-factorized Gaussian approximation (right).
    However, each divergence returns a different estimate of variance.
    For the $\alpha$-divergence, we use $\alpha=1/2$.
    }}
    \label{fig:ellipse_mf_t}
\end{figure}

\Cref{thm:even-symmetry} allows for several misspecifications in the variational approximation.
For example, $q$ may be factorized while $p$ is not, or $q$ and $p$ may behave differently in their tails. \Cref{fig:ellipse_mf_t} (left) illustrates VI's ability to recover the mean
when $p(z)$ is a 2-dimensional Student-t distribution with 5~degrees of freedom and correlation $\rho\! =\! 0.7$, and $\mathcal Q$ is a family of factorized Gaussians. 
For this target $p$ and variational family $\mathcal{Q}$, we recover the same, exact estimate of the mean~$\nu$ upon minimizing (via grid search) any of the $f$-divergences in \Cref{tab:divergences}, including the total variation distance.

In \Cref{app:asymmetric}, we provide an illustrative example of a skewed and therefore \textit{asymmetric} target, showing that in the presence of asymmetry, VI no longer recovers the mean.
The error in the mean estimate depends both of the amount of skewness and the choice of divergence.

\subsection{Guarantees from elliptical symmetry}
\label{sec:elliptical}

Next we present guarantees for the optimal scale of the variational approximation.
In our next theorem, we make the following regularity assumptions.
First, we take $f$ to be strictly convex and $p$ and $q$ to be strictly unimodal, as we did in \Cref{thm:even-symmetry}.
Next, we require $g(r)r^d$ to be monotone decreasing in the tails, as defined below.

\begin{bluebox}
    \begin{definition}[{\bf Radially decreasing in tail}]
        We say $g_p$ is radially decreasing in its tail if there exists some $R\!>\!0$ such that $g_p(r) r^d$ is decreasing for all for $r\!>\!R$.
    \end{definition}
\end{bluebox}
One can verify that the above condition holds for all standard elliptical distributions (\Cref{tab:elliptical-distribution}), including heavy-tail distributions.

\begin{bluebox}
\begin{theorem}[{\bf Recovery of elliptical symmetry}] \label{thm:elliptical}
Let $D_f$ be an $f$-divergence with $f$ strictly convex, let $p$ be an elliptically symmetric distribution with point of even symmetry $\mu$ and scatter matrix $M$, and let $\mathcal Q$ be an elliptical location-scale family with location parameter $\nu$ and scale parameter $S$.
In addition, suppose that $p$ is strictly unimodal and its radial function $g_p$ is differentiable everywhere and strictly radially decreasing in its tail.
Finally, suppose that any $q \in \mathcal Q$ is strictly unimodal and has a radial function $g_q$ which is everywhere continuous.

Then any stationary point of $D_f(p||q)$ with respect to the location-scale parameters $(\nu, S)$ of $\mathcal Q$ occurs at $\nu\! =\! \mu$ and $S\! =\! \gamma^2 M$ for some $\gamma > 0$. 
\end{theorem}
\end{bluebox}

The first equality, $\nu = \mu$, follows from \Cref{thm:even-symmetry}.
This result, combined with \Cref{lemma:congruence} and \cref{eq:S-dependence} in the proof of \Cref{lemma:stationary}, establishes that the problem of minimizing $D_f(p||q)$ over $S$ can be reduced to an optimization problem over the diagonal matrix $\Gamma$.
%
The proof of \Cref{thm:elliptical} is then broken up into two lemmas.
The first lemma establishes that stationary points of $D_f(p||q)$ can only occur where $\Gamma$ is 
proportional to the identity matrix, which per \Cref{lemma:stationary} implies $S \propto M$.
The second lemma establishes the existence of at least one such stationary point.

To prove the first lemma, we assume that $f$ is strictly convex and that $p$ and $q$ are strictly unimodal.

\begin{bluebox}
\begin{lemma}[{\bf All stationary points of $D_f$ recover the scatter matrix}] \label[lemma]{lemma:gamma}
Let $D_f$ be an $f$-divergence with $f$ strictly convex, let $p$ be an elliptically symmetric distribution with scatter matrix $M$ and let $\mathcal Q$ be an elliptical location-scale family with scale parameter $S$.
Assume further that for any $q \in \mathcal Q$, $q$ and $p$ have the same point of even symmetry.
Next, suppose that $p$ is strictly unimodal.

Then, any stationary point of $D_f(p||q)$ with respect to $\Gamma$, as defined in \cref{eq:Df_gl} must be such that $\Gamma = \gamma I$ for some $\gamma \in \mathbb R$.
This in turn implies that $S = \gamma^2 M$.
\end{lemma}
\end{bluebox}
\begin{proof}
    We denote the diagonal elements of $\Gamma$ by $\text{Diag}(\Gamma) = (\gamma_1, \gamma_2, \cdots, \gamma_d)$.
    Since $q$ and~$p$ share the same point of even symmetry, their $f$-divergence is given by \cref{eq:Df_gl} evaluated at~\mbox{$\tau\!=\!0$}. Stationary points occur where all the partial derivatives of this $f$-divergence vanish. Computing $\partial D_f(p||q)/\partial\gamma_i$ from \cref{eq:Df_gl}, we find that it is equal to zero when
  %
%
 %
 \begin{align} \label{eq:Gamma-stationary}
  \int f' \left( \frac{g_p(||\Gamma z||)}{g_q(||z||)} |\Gamma| \right) \left[ g_p(||\Gamma z||) + g_p'(||\Gamma z||) \dfrac{z_i^2 \gamma_i^2}{||\Gamma z||} \right]  \text d z & = 0.
\end{align}
Notice that only the second term in the sum inside the bracket depends on the index $i$.
Hence, in order to cancel out with the first term, the second term cannot depend on $i$ after integration.
This implies that,
\begin{equation} \label{eq:index-equality}
    \int f' \left( \frac{g_p(||\Gamma z||)}{g_q(||z||)} |\Gamma| \right) g_p'(||\Gamma z||) \dfrac{z_i^2 \gamma_i^2}{||\Gamma z||} \text d z = \int f' \left( \frac{g_p(||\Gamma z||)}{g_q(||z||)} |\Gamma| \right) g_p'(||\Gamma z||) \dfrac{z_j^2 \gamma_j^2}{||\Gamma z||} \text d z,
\end{equation}
for any pair $i, j$.
To proceed, let $\omega=\Gamma z$,
and as shorthand, let~$\Psi$ denote
the difference between the left and right sides of \cref{eq:index-equality}.
Then, 
%
\begin{equation}
    \Psi = \int f' \left( \frac{g_p(||\omega||)}{g_q(||\Gamma^{-1} \omega||)} |\Gamma| \right) \frac{g_p'(||\omega||)}{||\omega||} (\omega_i^2 - \omega_j^2) \frac{1}{|\Gamma|} \text d \omega.
    \label{eq:psi}
\end{equation}
%

Next, we observe that many of the terms in the integrand of \cref{eq:psi} (in particular, those involving the norm $\|\omega\|$ and the determinant $|\Gamma|$) are invariant under exchange of the coordinates $\omega_i$ and~$\omega_j$. Not all of the terms in the integrand, however, are invariant: the non-invariant terms are those involving the norm $\|\Gamma^{-1}\omega\|$ and the difference $\omega_i^2\!-\!\omega_j^2$. With these observations in mind, we define the set 
$$\Omega^+ = \{\omega : \omega_i^2>\omega_j^2\},$$ 
which accounts for exactly half the volume of $\mathbb{R}^d$. Then accounting for the non-invariant terms in the integrand, and noting that the integrand vanishes at the boundary of $\Omega^+$ (where $\omega_i^2=\omega_j^2)$, we can rewrite the integral over all of~$\mathbb{R}^d$ in \cref{eq:psi} as the \textit{folded} integral 
over~$\Omega^+$,
\begin{equation}
    \Psi = \int_{\Omega^+} \left [f' \left( \frac{g(||\omega||)}{g_q(||\Gamma^{-1} \omega||)} |\Gamma| \right) - f' \left( \frac{g(||\omega||)}{g_q(||\tilde \Gamma^{-1} \omega||)} |\Gamma| \right) \right] \frac{g'(||\omega||)}{||\omega||} (\omega_i^2 - \omega_j^2) \frac{1}{|\Gamma|} \text d \omega,
\end{equation}
where $\tilde \Gamma$ is obtained by interchanging $\gamma_i$ and $\gamma_j$ in $\Gamma$.
With this expression, we can complete the proof by contradiction.
Assume $\gamma_i \neq \gamma_j$ and without loss of generality, suppose $\gamma_i < \gamma_j$.
Under these conditions, we show that the integrand is nonnegative on all of $\Omega^+$ and strictly positive on some measurable set.

By assumption, $p$ is strictly unimodal and so $g_p'(||\omega||) < 0$ for $||\omega||>0$, that is for all $\omega \in \Omega^+$.
Therefore, the terms outside the bracket in the integrand are strictly negative.
Next,
\begin{equation}
    ||\Gamma^{-1} \omega|| = \sqrt{\frac{\omega^2_i}{\gamma^2_i} + \frac{\omega^2_j}{\gamma^2_j} + \sum_{k\neq i,j} \frac{\omega_k^2}{\gamma^2_k}}\ >\, \sqrt{\frac{\omega^2_j}{\gamma^2_i} + \frac{\omega^2_i}{\gamma^2_j} + \sum_{k\neq i,j} \frac{\omega_k^2}{\gamma^2_k}} = ||\tilde \Gamma^{-1} \omega||,
\end{equation}
where the strict inequality follows from $\omega^2_i > \omega^2_j$ and $\gamma^2_i < \gamma^2_j$.
By assumption of strict unimodality, $g_q$ is strictly decreasing and so $g_q(||\Gamma^{-1} \omega||) < g_q(||\tilde \Gamma^{-1} \omega||)$.
By strict convexity of $f$, $f'$ is strictly increasing.
Therefore,
\begin{equation}
    f' \left( \frac{g_p(||\omega||)}{g_q(||\Gamma^{-1} \omega||)} |\Gamma| \right) > f' \left( \frac{g_p(||\omega||)}{g_q(||\tilde \Gamma^{-1} \omega||)} |\Gamma| \right),
\end{equation}
and $\Psi < 0$.
This \textit{contradicts} \cref{eq:index-equality}.
Therefore, there is no stationary point such that $\gamma_i \neq \gamma_j$ for any pair $i, j$.
Moreover, to find a stationary point, we require that $\Gamma = \gamma I$ for some $\gamma \in \mathbb R$.
Finally, if $\Gamma = \gamma I$, it follows from \Cref{lemma:stationary} that $S = \gamma^2 M$.
\end{proof}

The next lemma shows that there exists a stationary point for $\gamma > 0$.
This result requires some further technical assumptions.
First, we require that $g_q$ is continuous everywhere.
Second, we require that $g_p(r)r^d$ is strictly monotone decreasing in its tails.


\begin{bluebox}
\begin{lemma}[{\bf Existence of a stationary point}]\label[lemma]{lemma:gamma-exists}
    Assume the setup in \Cref{lemma:gamma} and furthermore, suppose $g_q$ is everywhere continuous and strictly radially decreasing in its tail. 
    Then there exists a stationary point of $D_f(p||q)$ with respect to $\Gamma$, as defined in \cref{eq:Df_gl}, such that $\Gamma = \gamma I$ for some $\gamma\!>\!0$.
\end{lemma}
\end{bluebox}

The proof of \Cref{lemma:gamma-exists} is in \Cref{app:radial-tail}.
The idea is to show that $\partial D_f(p||q) / \partial \gamma$ or equivalently the right hand of \cref{eq:Gamma-stationary}, has opposite signs in the limits where $\gamma\!\to\!0$ and $\gamma\!\to\!+\infty$, and so by continuity, there must exist at least one value of $\gamma$ such that $\partial D_f(p||q) / \partial \gamma\!=\!0$.

The proof of \Cref{thm:elliptical} follows from \Cref{thm:even-symmetry} and \Cref{lemma:congruence,lemma:gamma,lemma:gamma-exists}.
An immediate consequence of \Cref{thm:elliptical} is that if $p$ admits a second moment, then VI recovers the mean and correlation matrix of $p$.
\Cref{fig:ellipse_mf_t} (right) illustrates VI's ability to recover these statistics when $p$ is a Student-t distribution and $\mathcal{Q}$ is the family of multivariate Gaussians.

\subsection{Guarantees from partial symmetry}
\label{sec:partial}

We now provide guarantees when $p$ exhibits partial symmetry.
We use $\sigma$ to denote the coordinates along which $p$ is symmetric and $\bar \sigma$ to denote the remaining coordinates.
In this notation, we decompose the variational parameters of $q$ as
\begin{equation}
  \nu = \begin{pmatrix} \nu_\sigma \\ \nu_{\bar \sigma} \end{pmatrix}, \ \
  S = \begin{pmatrix} S_{\sigma \sigma} & S_{\sigma \bar \sigma} \\ S^T_{\sigma \bar \sigma} & S_{\bar \sigma \bar \sigma} \end{pmatrix},
\end{equation}
and we use $\nu_{\sigma|\bsigma}$ to denote the conditional mean of $q$ and $S_{\sigma | \bsigma}$ its conditional scale matrix.
If $q(z)$ is elliptically symmetric, we also have the following relationships \citep[Theorem 2.18]{Fang:1990}, 
%
\begin{align}
     \label{eq:cond-mean} 
     \nu_{\sigma|\bsigma} &= \nu_\sigma + S_{\sigma \bar \sigma} S^{-1}_{\bar \sigma \bar \sigma} (\zb - \nu_{\bar \sigma}), \\
    \label{eq:cond-scale}
      S_{\sigma|\bsigma} & \propto (S_{\sigma \sigma} -  S_{\sigma \bar \sigma} S^{-1}_{\bar \sigma \bar \sigma} S_{\bar \sigma \sigma}).
\end{align}
Note that while the conditional scatter matrix of an elliptical distribution is constant with respect to $\zb$, its covariance matrix is not (in general).
Rather, $\text{Cov}_q(\zs|\zb) = k(\zb) S_{\sigma|\bsigma}$ for some scalar function $k$.
See \citet{Fang:1990} for more discussion.
We similarly decompose the mean $\mu = (\mu_\sigma, \mu_{\bar \sigma})$ of the target density $p$ and denote $M_{\sigma|\bsigma}$ and $M_{\sigma \sigma}$ its conditional and marginal scatter matrix respectively.

\begin{table}
\begin{center}
\begin{tabular}{r l l}
\rowcolor{Cerulean!10} {\bf Divergence} & $h$ & $K$ \\
$\text{KL}(q||p)$ & $q(\zb)$ & $\text{KL}(q(\zb) || p(\zb))$ \\
\rowcolor{Cerulean!10} $\text{KL}(p||q)$ & $p(\zb)$ & $\text{KL}(p(\zb) || q(\zb))$ \\
$D_\alpha(p||q)$ & $p(\zb)^\alpha q(\zb)^{1-\alpha}$ & 0
\end{tabular}
\caption{\textit{Example of $f$-divergences which verify the chain rule (\Cref{def:chain-rule})}} \label{tab:separable}
\end{center}
\end{table}

We can readily leverage our work in the previous sections by setting a few conditions.
First, we assume $D_f(p||q)$ verifies a chain rule, as defined below.
\begin{bluebox}
\begin{definition} [\bf Chain rule of $f$-divergence] 
\label[definition]{def:chain-rule}
  We say an $f$-divergence verifies a chain rule if there exists functions $h$ and $K$ such that,
  \begin{equation} \label{eq:sigma-separable}
    D_f [p(z) || q(z)] = \int_{\bar \sigma} D_f \big[p(\zs | \zb) || q(\zs | \zb) \big] \ h \big( p(\zb), q(\zb) \big) d \zb + K \big( p(\zb), q(\zb) \big),
  \end{equation}
  and $h \big( p(\zb), q(\zb) \big) > 0$.
  Note that $h$ is a function of the marginal densities of $p$ and $q$ evaluated at $\zb$, while  $K$ is a function of the marginal distributions of $p$ and $q$ but not $\zb$ itself.
\end{definition}
\end{bluebox}
The existence of a chain rule is not particularly constraining and is, for instance, verified by both KL-divergences and all $\alpha$-divergences; see \Cref{tab:separable}. 


Next, we assume that $p(\zs|\zb)$ is elliptically symmetric and that its conditional mean $\mu_{\zs|\zb}$ depends linearly in $\zb$, while its conditional scatter matrix $M_{\zs|\zb}$ is constant. 
(Once again, this does not exclude the possibility that the conditional covariance depends on $\zb$.)
These restrictions are verified by the conditional distributions of elliptical distributions and also non-trivial densities such as the Rosenbrock distribution~\citep{Roberts:1997}
and Neal's funnel from \cref{eq:funnel}.
The latter often characterizes hierarchical priors in Bayesian models.

To state the theorem, we introduce (as before) the radial functions, $g_p(\cdot |\zb)$ and $g_q(\cdot|\zb)$,
this time defined in terms of conditional distributions.
\begin{bluebox}
\begin{theorem}[\bf Recovery of the partial elliptical symmetry along~$\sigma$] \label{thm:elliptical-sigma}
Let $D_f$ be an $f$-divergence which verifies the chain rule (\Cref{def:chain-rule}) and with $f$ strictly convex; let $p$ be elliptically symmetric along $\sigma$ with a point of even symmetry $\mu_{\sigma|\bsigma}$ that varies linearly with $\zb$ and a constant scatter matrix $M_{\sigma|\bsigma}$; and let $\mathcal Q$ be an elliptical location-scale family with location parameter $\nu$ and scale parameter $S$.
Also, suppose $g_p(\cdot |\zb)$, as defined above, is differentiable everywhere, strictly unimodal, and strictly radially decreasing in its tail.

Then any  minimizer of $D_f(p||q)$ satisfies $\nu_{\sigma|\bsigma} = \mu_{\sigma|\bsigma}$ and $S_{\sigma | \bsigma} = \gamma^2 M_{\sigma | \bsigma}$ for some $\gamma > 0$.
\end{theorem}
\end{bluebox}

\begin{proof}
  The keystone of the proof is to find values for $(\nu_\sigma,S_{\sigma\sigma},S_{\sigma\bar\sigma})$ that  minimize,
  \begin{equation} \label{eq:conditional-divergence}
      \int D_f \big [p(\zs|\zb)||q(\zs|\zb) \big] h \big( p(\zb), q(\zb) \big) \text d \zb,
  \end{equation}
  and then to note that this minimizer is found for any value of $(\nu_{\bar\sigma},S_{\bar\sigma\bar\sigma})$ that fix $q(\zb)$ and hence leave the remaining terms in $D_f(p||q)$ (\cref{eq:sigma-separable}) unchanged.

  From \Cref{thm:even-symmetry}, the conditional divergence $D_f[p(\zs|\zb)||q(\zs|\zb)]$ is minimized by matching the mean.
  By assumption, $\mu_{\sigma|\bsigma} = \alpha + A\zb$,
  for some $\alpha \in \mathbb R^{|\bsigma|}$ and $A \in \mathbb R^{|\bsigma|\times|\sigma|}$.
  Since $q$ is elliptically symmetric, $\nu_{\sigma|\bsigma}$ is also linear in $\zb$ and moreover, we require,
  \begin{equation}
      \nu_{\sigma|\bsigma} = \nu_\sigma + S_{\sigma \bar \sigma} S^{-1}_{\bar \sigma \bar \sigma} (\zb - \nu_{\bar \sigma}) = \alpha + A \zb,
  \end{equation}
  which is solved by,
  \begin{equation} \label{eq:conditional-mean}
      \nu_\sigma = \alpha + A \nu_{\bsigma}, \ \ S_{\sigma \bsigma} = A S_{\bsigma \bsigma}.
  \end{equation}
  Then, at each $\zb$, the conditional divergence in the integral is minimized with respect to the variational mean of $q(\zs|\zb)$.
  
  It remains to find $S_{\sigma \sigma}$.
  \Cref{lemma:conditional-scatter} in \Cref{app:elliptical-sigma} adapts \Cref{thm:elliptical} and shows that \cref{eq:conditional-divergence} would be minimized with respect to $q(\zs|\zb)$ if $S_{\sigma|\bsigma} = \gamma^2 M_{\sigma|\bsigma}$ for some $\gamma > 0$.
  Now from \cref{eq:cond-scale}, we have that $S_{\sigma|\bsigma} = \beta (S_{\sigma \sigma} - S_{\sigma \bsigma} S^{-1}_{\bsigma \bsigma} S_{\bsigma \sigma})$ for some $\beta > 0$.
  Then, a minimizer for $q(\zs|\zb)$ is found by setting,
  \begin{equation} \label{eq:conditonal-scatter}
      S_{\sigma \sigma} = \beta^{-1} \gamma^2 M_{\sigma|\bsigma} + S_{\sigma \bsigma} S^{-1}_{\bsigma \bsigma} S_{\bsigma \sigma}.
  \end{equation}
  (Note that $\gamma$ in the above expression is chosen to minimize the average conditional divergence $\int D_f[p(\zs|\zb)||q(\zs|\zb)] h(\zb) \text d \zb$ but does not minimize the pointwise conditional divergence for each $\zb$.)
  Hence, \cref{eq:conditional-mean,eq:conditonal-scatter} can be solved for any choice of $\nu_{\bsigma}$ and $S_{\bsigma \bsigma}$.
\end{proof}

If, in addition, we assume that the conditional point of even symmetry of $p$ is constant with respect to $\zb$, then we obtain guarantees for the recovery of the marginal elliptical symmetry.
This scenario arises in the elliptical funnel, the Rosenbrock distribution and the product of independent elliptical location-scale distributions.


\begin{bluebox}
\begin{corollary}[\bf Recovery of the marginal mean and scatter matrix along $\sigma$] \label[corollary]{cor:elliptical-sigma}
 Assume the same conditions as in \Cref{thm:elliptical-sigma} and assume further that the conditional point of even symmetry of $p$ along $\sigma$ is constant with respect to $\zb$.
 Then any minimizer of $D_f(p||q)$ satisfies $\nu_\sigma = \mu_\sigma$ and $S_{\sigma \sigma} = \gamma^2 M_{\sigma \sigma}$ for some $\gamma > 0$.
\end{corollary}
\end{bluebox}

\begin{proof}
    Recall $\mu_{\sigma|\bsigma} = \alpha + A\zb$.
    By assumption, $\mu_{\sigma|\bsigma}$ is constant with respect to $\zb$ and so $A = \bf 0$.
    Then from \cref{eq:conditional-mean} and \cref{eq:conditonal-scatter}, we have $\nu_\sigma = \mu_{\sigma|\bsigma}$ and $S_{\sigma \sigma} = \gamma^2 M_{\sigma|\bsigma}$.
    Finally, in \Cref{lemma:constant-symmetry} of \Cref{app:elliptical-sigma}, we show that $\mu_{\sigma|\bsigma}=\mu_\sigma$ and $M_{\sigma \sigma} \propto M_{\sigma|\bsigma}$ if $p$ is elliptically symmetric along $\sigma$ with a constant point of even symmetry and a constant scatter matrix.
\end{proof}

In \Cref{app:funnel}, we apply these results to the funnel of \cref{eq:funnel} and variations thereof.

\subsection{Related guarantees} \label{sec:related-guarantees}

We now examine related guarantees which have appeared previously in \citet{Margossian:2025robust} and concurrently in \citet{Zellinger:2026}.

We relax the assumption of log-concavity on $p$ made by \citet{Margossian:2025robust}, and instead we require $p$ and $q$ to be unimodal \textit{and} elliptically symmetric. Under these conditions, we show that $q$ matches the symmetry of $p$ and recovers its mean and correlation matrix if $p$ has finite second moments.
%
These results are not a strict generalization of the mean-recovering result by \citet[Theorem 8]{Margossian:2025robust}, where guarantees to recover the mean require log-concavity and even symmetry but not elliptical symmetry.
An example of a log-concave target which is even-symmetric but not elliptically symmetric is a product of univariate Laplace distributions.\footnote{A product of univariate Laplace distributions does not yield 
a multivariate Laplace distribution; the latter is elliptically symmetric, while the former is not. Likewise a product of univariate Student-t distributions does not yield a multivariate Student-t distribution.} 
For such a target, however, we still obtain guarantees for the exact recovery of the mean from our results on partial symmetry (\Cref{cor:elliptical-sigma}), because any even-symmetric univariate distribution is trivially elliptically symmetric.
Our guarantees can further handle the similar case where $p$ is a product of univariate Student-t,  which is no longer log-concave. 

Independently of our work, \citet{Zellinger:2026} studied conditions under which the minimizers of the forward KL and certain $\alpha$-divergences recover the symmetry and thence the mean and the correlation matrix.
Their results are stronger than ours in some respects and weaker in others. 
On one hand, they obtain new guarantees for VI when $p$ is heavy-tailed or even multimodal.
On the other hand, they obtain these guarantees by replacing the condition of log-concavity on $p$ by a strong convexity condition on $q$.
In particular, when minimizing the forward KL, they require that
the base distribution $q_0$ of the approximating location family must be log-concave, and when
minimizing an $\alpha$-divergence, they require that $q_0^{1-\alpha} / (\alpha(1\!-\!\alpha))$ must be convex.
%
While conditions on $q_0$ are generally preferable to conditions on $p$ (since the practitioner chooses $q_0$), we note that these conditions are difficult to meet for the most common settings of VI. For example, if $q_0$ is a standard normal distribution,
the condition of convexity is only verified for $\alpha > 1$.
Moreover, these guarantees mostly apply to VI with the forward KL or an $\alpha$-divergence with $\alpha>1$,
divergences which are notoriously difficult to optimize.
Unlike our results, these guarantees do not readily apply to the reverse-KL, the more approachable $\alpha$-divergences with $\alpha \in (0, 1)$ and other $f$-divergences.

One may wonder whether the specificity of these guarantees reveal 
a gap in our 
current theoretical understanding of VI. However, examples can be constructed when $p$ is multimodal and VI does not match the even symmetry of the target.
Indeed, when minimizing the reverse-KL, VI is known to collapse to one mode, rather than match the even symmetry \citep[e.g.,][]{Turner:2011, Margossian:2025robust} and similar examples are constructed by \citet{Zellinger:2026} for $\alpha$-divergences of order $\alpha < 1$.

\section{Numerical Experiments} \label{sec:experiments}

\begin{table*}[]
    \centering
    \small
    \begin{tabular}{l r l r r r}
    & & & & (nc) & (marg) \\[-0.05in]
    \textbf{Call name} & $d$ & \textbf{Description} & $a_\text{90}$ & $a_\text{90}$ & $a_\text{90}$ \\
    \rowcolor{gray!20} \texttt{student} & 2 & Elliptical target with heavy tails. & 0 & & \\
    \texttt{crescent} & 3 & Elliptical Rosenbrock distribution. & 29.36 & & \\
    \rowcolor{gray!20} \texttt{funnel} & 4 & Elliptical funnel with partial symmetry (\cref{eq:funnel}). & 112.53 & & \\
    \texttt{schools} & 10 & Bayesian hierarchical model for education data.  & 285.36 & 3.57 & 3.27 \\
    \rowcolor{gray!20} \texttt{disease} & 102 & Gaussian process model for epidemiology data. & 231,840 & 137.26 & 2.22 \\
    \texttt{SKIM} & 305 & Sparse kernel interaction model for genetic data. & 220,493 & 154.41 & 785.33
    \end{tabular}
    \caption{\textit{Target distributions for experiments and estimated even-asymmetry $a_\text{90}$; see \cref{eq:symmetry}.
    The hierarchical models (\texttt{schools}, \texttt{disease}, \texttt{SKIM}) were implemented using (i) a standard centered parameterization, (ii) a non-centered parameterization with less asymmetry, (iii) an approximate marginalization of the latent variables, yielding a collapsed (and even less asymmetric) posterior; see the discussion in \Cref{sec:sym}.}}
    \label{tab:targets}
\end{table*}

We investigate the performance of VI on a diverse set of target densities (\Cref{tab:targets}).
The first three are synthetic---a multivariate Student-t, an elliptical funnel, and a crescent distribution---and the others are derived from Bayesian hierarchical models, including a model of education data~\citep{Gelman:2013}, a sparse kernel interaction model of gene microarray data~\citep{Agrawal:2019}, and a Gaussian process model of mortality counts~\citep{Vanhatalo:2019}.
The full definition of each model is provided in \Cref{app:targets} and code to reproduce the experiments can be found at \url{https://github.com/charlesm93/VI_elliptical_symmetry}.

\subsection{Are Bayesian posteriors symmetric?}
\label{sec:sym}

We now discuss how approximate symmetries may arise in a Bayesian analysis.
Given a \mbox{latent variable~$z$} and observation $x$, the posterior is given by $\pi(z|x)\!\propto\! \pi(z)\, \pi(x|z)$,
and the likelihood often factors as $\pi(x|z)\! =\! \prod_i \pi(x_i|z)$.
In general this product may not have an even or elliptical symmetry even when one is found in each of its terms.
However, if the product is dominated by a symmetric prior (for sparse data) or the likelihood (for rich data), then the posterior tends to be approximately symmetric~\citep{Margossian:2025robust}. In the latter regime, the Bernstein-von Mises theorem \citep{Vaart:1998} is often invoked to argue that the posterior is approximately Gaussian. 

One expects less symmetry in hierarchical models with asymmetric priors, as in \cref{eq:funnel}.
Further complexity arises when the prior mean and the correlation $C$ for $\theta$ depend on additional hyperparameters \mbox{$\mu, \rho \in \mathbb R$} 
%
  and $\theta \sim \mathcal N(\mu, e^{2 \tau} C(\rho))$,
%
as in Gaussian processes and, for example, the \texttt{disease} and \texttt{SKIM} targets in \Cref{tab:targets}.
But there are also strategies to mitigate these sources of asymmetry, without changing the generative model. 
One strategy is to use a \textit{non-centered parameterization} \citep{Papaspiliopoulos:2007}. Let $L$ denote the Cholesky decomposition of the prior covariance matrix, such that \mbox{$LL^T\! =\! \exp(2 \tau) C(\rho)$}. This strategy introduces 
an auxiliary variable \mbox{$\varepsilon\!\sim\!\mathcal{N}(0,1)$} and 
recomputes the likelihood as 
\mbox{$\pi(x|\varepsilon) = \pi(x|\theta\! =\! L \varepsilon + \mu)$}.
A Bayesian inference algorithm then approximates the posterior \mbox{$\pi(\varepsilon, \mu, \tau, \rho| x)$}, where the funnels
in the prior have been removed due to the independence of $\varepsilon$ and~$\tau$.
A caveat is that this non-centered parameterization can complicate the likelihood, especially in rich data regimes, leading to a challenging posterior geometry and it is often unclear in practice which of the centered or non-centered parameterization produces a better-behaved posterior geometry.

A second strategy is to marginalize out $\theta$ and perform inference over the collapsed posterior,
$
\pi(\mu, \tau, \rho| x) \propto  \pi(\mu, \tau, \rho) \int_\Theta \pi(x, \theta |\mu, \tau, \rho)\, \text d \theta.
$
Here the marginalization serves to remove the funnel over $\theta$ and $\tau$ in the prior. 
Inference on $\theta$ is performed post-hoc by approximating \mbox{$\pi(\theta|\mu, \tau, \rho, x)$}.
If the likelihood is normal, then the marginalization is tractable; otherwise it must be approximated---for example, via an integrated Laplace approximation \citep[e.g.,][]{Rasmussen:2006, Rue:2009}.

We implement each hierarchical model 
in three ways, via a standard (centered) parameterization, a non-centered parameterization, and a marginalized target (where the marginalization is performed exactly for \texttt{schools} and approximately for \texttt{dissease} and \texttt{SKIM}).
Empirically, we find that the latter strategies produce less asymmetric targets; see \Cref{tab:targets}.

\subsection{VI algorithm}

We use automatic differentiation VI \citep[ADVI,][]{Kucukelbir:2017}, as implemented in {\sc Stan} \citep{Carpenter:2017, Stan:2025}.
ADVI employs a Gaussian approximation 
and minimizes $\text{KL}(q||p)$ via stochastic optimization.
Details on the implementation of ADVI are described in \Cref{app:algorithm}.
While our theorems consider more general $f$-divergences, we do not have a reliable way to minimize these divergences on the range of considered problems, and so we only give results for $\text{KL}(q||p)$.
For models with a collapsed posterior, we use {\sc Stan}'s prototype integrated Laplace approximation \citep{Margossian:2020}.
As a benchmark, we approximate $\pi(z|x)$ with exact sampling of the synthetic targets and long runs (20,000 iterations 
) 
of {\sc Stan}'s adaptive Hamiltonian Monte Carlo sampler \citep{Hoffman:2014, Betancourt:2018} with a non-centered parameterization for the hierarchical models, thereby obtaining golden benchmark estimates of the posterior mean, standard deviation and correlations.

\subsection{Results}

\begin{figure}
    \centering
    \includegraphics[width=0.8\linewidth]{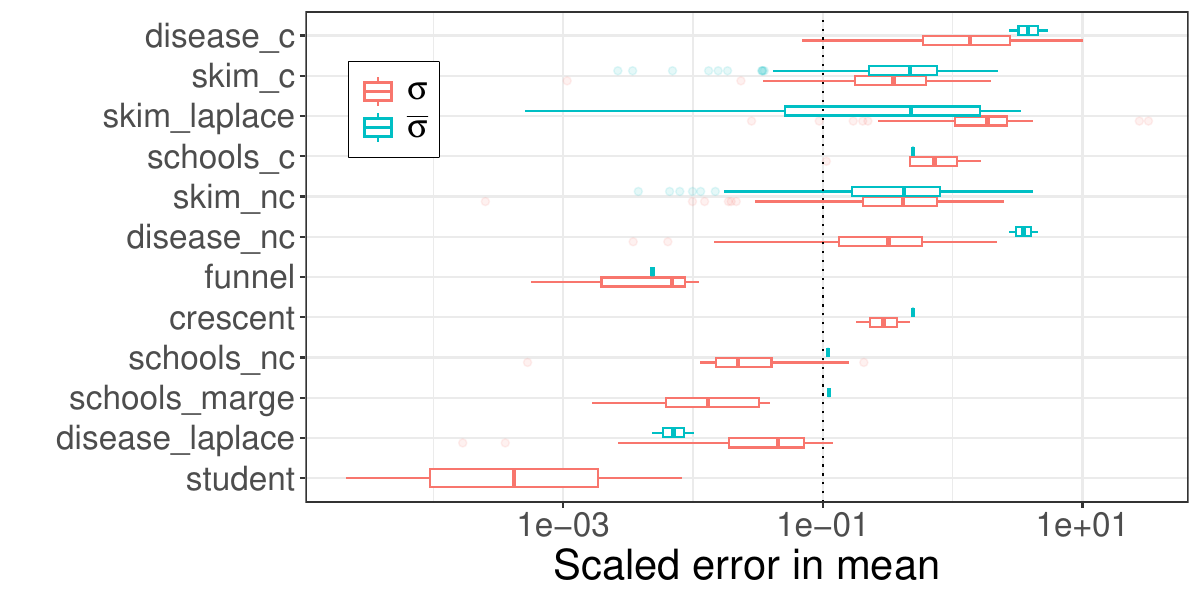}
    \caption{\textit{Absolute errors in VI's mean estimate scaled by the target standard deviation along all coordinates.
    The targets are ordered, bottom to top, from most to least even-symmetric.
    For targets with partially symmetric priors, we distinguish coordinates $\sigma$ which are a priori symmetric from non-symmetric coordinates $\bar \sigma$.
    As a trend, VI returns better estimates of the mean for more even-symmetric targets.
    The mean is also better estimated in \texttt{crescent} and \texttt{disease} along the coordinates~$\sigma$.
    The dotted line is the Monte Carlo standard error obtained with 100 independent posterior draws.
    }}
    \label{fig:experiment}
\end{figure}

Our first results investigate the ability of VI to recover the mean in the presence (or absence) of even-symmetry. For each target density in \Cref{tab:targets}, we stochastically measure its even-asymmetry. 
Given a sample $z$ from $p(z)$, we compute a reflected sample $z' = \hat \mu - z$, where~$\hat \mu$ is the benchmark estimate of the mean of $p$. Then we measure the asymmetry of the target by computing
\begin{equation}
 \label{eq:symmetry}
    a(z) = |\log \pi(z, x) - \log \pi(z', x)|.
\end{equation}
If $\pi(x, z)$ is even-symmetric about $\hat \mu$, then $a(z)\! =\! 0$ for all $z$.
We evaluate $a(z)$ for 20,000 samples, and report its 90$^\text{th}$ quantile in \Cref{tab:targets}.
This procedure works for all of the models except \texttt{SKIM}, where there is numerical instability in the reflected density due to the integrated Laplace approximation (in  $\sim$7.5\% of cases); in this case, we report the $90^\text{th}$ quantile based only on samples for which $a(z)$ does not evaluate to \texttt{NaN}.

From the above procedure, we know which targets are approximately even-symmetric. 
Next we report the absolute error in the means 
that are estimated by VI across all coordinates of $z$,
scaled by the target's standard deviation.
We plot the distribution of scaled errors (mean and quantiles) in \Cref{fig:experiment}.
To contextualize this error, we indicate with a dotted line at $10^{-1}$ the Monte Carlo standard error, scaled by the standard deviation, we would obtain with 100 independent draws from the target $p$ or equivalently MCMC with an effective sample size of 100, which is often a reasonable standard error to aim for with MCMC \citep[e.g.,][]{Gelman:2011}, although the target error should of course be adjusted to the problem at hand.
Overall, we find that for more even-symmetric targets, VI yields better estimates of the mean.
We also plot the errors separately along coordinates which are a priori even-symmetric versus those which are not.
In the \texttt{crescent} and \texttt{disease} models, the mean is better estimated along the former, but in other models the errors are comparable.
The elliptical \texttt{funnel} presents an outlier: despite the target's relatively large asymmetry, VI accurately recovers the mean, including along the coordinate with no partial symmetry.
This does not contradict our theory but suggests that the absence of symmetry does not systematically lead to poor estimates of the mean.


We find a similar trend for estimates of correlations.
Generally, VI yields better estimates in models with more elliptical symmetry and along coordinates of partial elliptical symmetry. This trend is clear for the synthetic targets and \texttt{schools}, less so for \texttt{disease} and \texttt{SKIM}, perhaps because the latter have many nearly zero correlations.
We also examine estimates of the covariances and find that the covariance matrix may be poorly estimated even when the correlation matrix is accurately recovered.
We provide these results in \Cref{app:correlation}.

\section{Discussion}

In this paper, we extend several symmetry-based guarantees for VI. Our results hold broadly for targets that are unimodal but not necessarily log-concave, for models with partially symmetries in some but not all of their coordinates, and for optimizations involving any convex $f$-divergence, including but not limited to the reverse KL.
%

Our results show that in the presence of certain exact symmetries, VI can exactly recover certain corresponding statistics.
But it is more common in Bayesian modeling for the posterior to be nearly symmetric and for the statistics of the posterior to be approximately recovered.
A natural question is whether one can develop
error bounds based on a measure of asymmetry, such as 
the one we used 
in \Cref{sec:experiments}.
Such a direction for future research was 
proposed in earlier work \citep{Margossian:2025robust}.
As far as we can tell, the proof techniques we develop in this paper do not immediately extend to error bounds.
To pursue this line of work, we may draw inspiration from error bounds developed for VI in pre-asymptotic regimes where the target becomes approximately Gaussian \citep[e.g.,][]{Zhang:2020, Katsevich:2024}.

Our guarantees provide not only theoretical insight, but also prescriptions for practitioners using VI to approximate the posteriors of Bayesian hierarchical models. They suggest, in particular, that VI can be improved by implementing these models in certain ways, for example by (approximately) marginalizing out the latent variables $\theta$ or by using a non-centered parameterization.
These prescriptions are reminiscent of those known to improve the performance of MCMC on challenging posteriors \citep[e.g.,][]{Papaspiliopoulos:2007, Betancourt:2015, Gomez:2018, Margossian:2020}.

There exist other transformations to improve the posterior geometry and it would be of interest to study how these alter the symmetry of the posterior and in turn the performance of VI.
In the hierarchical models we examine, the data per latent variable $\theta$ is sparse and so the non-centered parameterization is a natural choice,
but in other applications (e.g., rich data or unbalanced data), we may obtain better results with a centered parameterization or a partially non-centered parameterization \citep{Papaspiliopoulos:2007}.
In a more general setting, the target can be made more even-symmetric using skewness reduction, an idea which was recently explored in the context of VI \citep{Chen:26}.

The above considerations suggest improving VI by transforming the target. But a more common approach is to choose a
richer variational family $\mathcal Q$---for example, one allowing skewed approximations~\citep{Tan:2024, Pozza:2026}, or even
semi-parametric or non-parametric approximations \citep[e.g.,][]{Agrawal:2020, Xu:2023, Cai:2024b, Xu:2025}.
While this approach requires fewer bespoke manipulations of $p$, its computational expense can grow quickly with the complexity of $\mathcal{Q}$. It is therefore of interest to understand how well simpler variational families can perform.
Moving forward, we advocate
a workflow
in which VI proceeds first with a restricted family $\mathcal Q$ (e.g., factorized or location-scale), the accuracy of the inference is checked (here, it would be interesting to develop a symmetry-based check), slight corrections are applied \citep[e.g.,][]{Yao:2018, Giordano:2018, Pozza:2026}
and $\mathcal Q$ is progressively complexified as necessary.

\section*{Acknowledgments}
We thank Brian Ward for help with {\sc BridgeStan} and Lena Zellinger, Antonio Vergari and Trevor Campbell for helpful discussion, as well as five anonymous reviewers for feedback on an early version of this manuscript.
We also thank three additional reviewers for their feedback on a more recent version of this manuscript.
%
CM acknowledges the support of the Natural Sciences and Engineering Research Council (NSERC) of Canada: RGPIN-2026-07219, DGECR-2026-00308.
IR is supported by an NSERC Graduate Research Scholarship (NSERC CGRS D).

\begin{singlespace}
\bibliography{ref.bib}
\end{singlespace}

\clearpage

\appendix

\section{Supporting Proofs and Illustrative Examples}

\subsection{Non-vanishing directional derivative for a solution which does not match the mean} \label[appendix]{app:proof-even-symmetry}

In this appendix, we complete the proof of \Cref{thm:even-symmetry} by showing that the directional derivative $D_f(p||q)\cdot v$ does not vanish when $v = \Gamma^{-2} \tau$ and $\tau \neq 0$ (meaning $q$ does not match the mean of $p$).

To alleviate the notation, let
\begin{align}
    t(\zeta) & = |\Gamma| \frac{g_p(||\Gamma(\zeta - \tau)||)}{g_q(||\zeta||)}. \label{eq:t} \\
    k(\zeta) & = |\Gamma| \frac{g_p'(||\Gamma(\zeta - \tau)||)}{||\Gamma(\tau-\zeta)||}
\end{align}
Here $t(\zeta)$ denotes the argument of $f'$ and $k(\zeta)$ the remaining radial terms in \cref{eq:gradient-u}. 
Note that $k(\zeta)$ is even-symmetric about $\tau$.
With this notation, we write the directional derivative as the sum of an integral over $\Omega^+$ and $\Omega^-$, 
\begin{align} \label{eq:omega-partition}
  \nabla_\tau D_f(p||q) \cdot v & = \int f' (t(\zeta)) k(\zeta)  (\tau - \zeta)^T\tau \ \text d \zeta \nonumber \\
  & = \int_{\Omega^+} f' (t(\zeta)) k(\zeta)  (\tau - \zeta)^T\tau \ \text d \zeta + \int_{\Omega^-} f' (t(\zeta)) k(\zeta)  (\tau-\zeta)^T\tau \ \text d \zeta,
\end{align}
with, by construction, the contribution of the integral over $\Omega^0$ vanishing.
We can ``fold'' this integral by using a transformation $T$, which returns the reflection of $\zeta$ about $\tau$, that is $T: \zeta \to \zeta'$ where $\zeta' = 2\tau - \zeta$.
Then, $(\tau-\zeta')^T \tau = (\tau - (2\tau - \zeta))^T \tau =  - (\tau- \zeta)^T \zeta$.
Hence, if $\zeta \in \Omega^+$, then $\zeta' \in \Omega^-$ and moreover $T(\Omega^+) = \Omega^-$.
Note further that $T(k(\zeta)) = k(\zeta)$, since $k(\zeta)$ is even-symmetric about $\tau$.
Applying $T$ to the second integral over $\Omega^-$, we obtain the \textit{folded} integral,
\begin{align}  \label{eq:omega-folded}
    \nabla_\tau D_f(p||q) \cdot v & = \int_{\Omega^+} f' (t(\zeta)) k(\zeta)  (\tau - \zeta)^T\tau \ \text d \zeta - \int_{\Omega^+} f' (t(\zeta')) k(\zeta) (\tau - \zeta)^T\tau \ \text d \zeta \nonumber \\
    & = \int_{\Omega^+} \left[f'(t(\zeta)) - f'(t(\zeta')) \right ] k(\zeta)  (\tau - \zeta)^T\tau \ \text d \zeta.
\end{align}
Over $\Omega^+$, $ (\tau - \zeta)^T\tau > 0$.
By assumption of $p$'s strict unimodality, $g_p'$ is strictly negative, except at $\zeta=\tau$, which lies on $\Omega^0$.
Therefore $k(\zeta)$ is strictly negative over $\Omega^+$.
We will now show that $f'(t(\zeta)) - f'(t(\zeta'))$ is also bounded by 0.
The norm of $\zeta'$ is,
\begin{equation}
    ||\zeta'||^2 = 4||\tau||^2 + ||\zeta||^2 - 4\tau^T||\zeta|| > ||\zeta||^2,
\end{equation}
where we obtain an inequality by noting that $\tau^T \zeta < ||\tau||^2$ for $\zeta \in \Omega^+$.
By strict unimodality, $g_q(||\zeta||) > g_q(||\zeta'||)$, and so we have ordered the denominator in $t(\zeta)$ (\cref{eq:t}).
The numerator, $g_p(||\Gamma(\zeta - \tau)||)$, is even-symmetric about $\tau$ 
and remains unchanged when evaluated at $\zeta$ or $\zeta'$.
Therefore, $t(\zeta) < t(\zeta')$.

Finally, we leverage the assumption that $f$ is strictly convex to obtain {$f'(t(\zeta)) - f'(t(\zeta')) < 0$}.
Thus, the integrand is strictly lower-bounded by 0, and the directional derivative is non-zero. 

\subsection{Mean estimate for asymmetric targets} \label[appendix]{app:asymmetric}

\begin{figure*}
    \centering
    \includegraphics[width=2.6in]{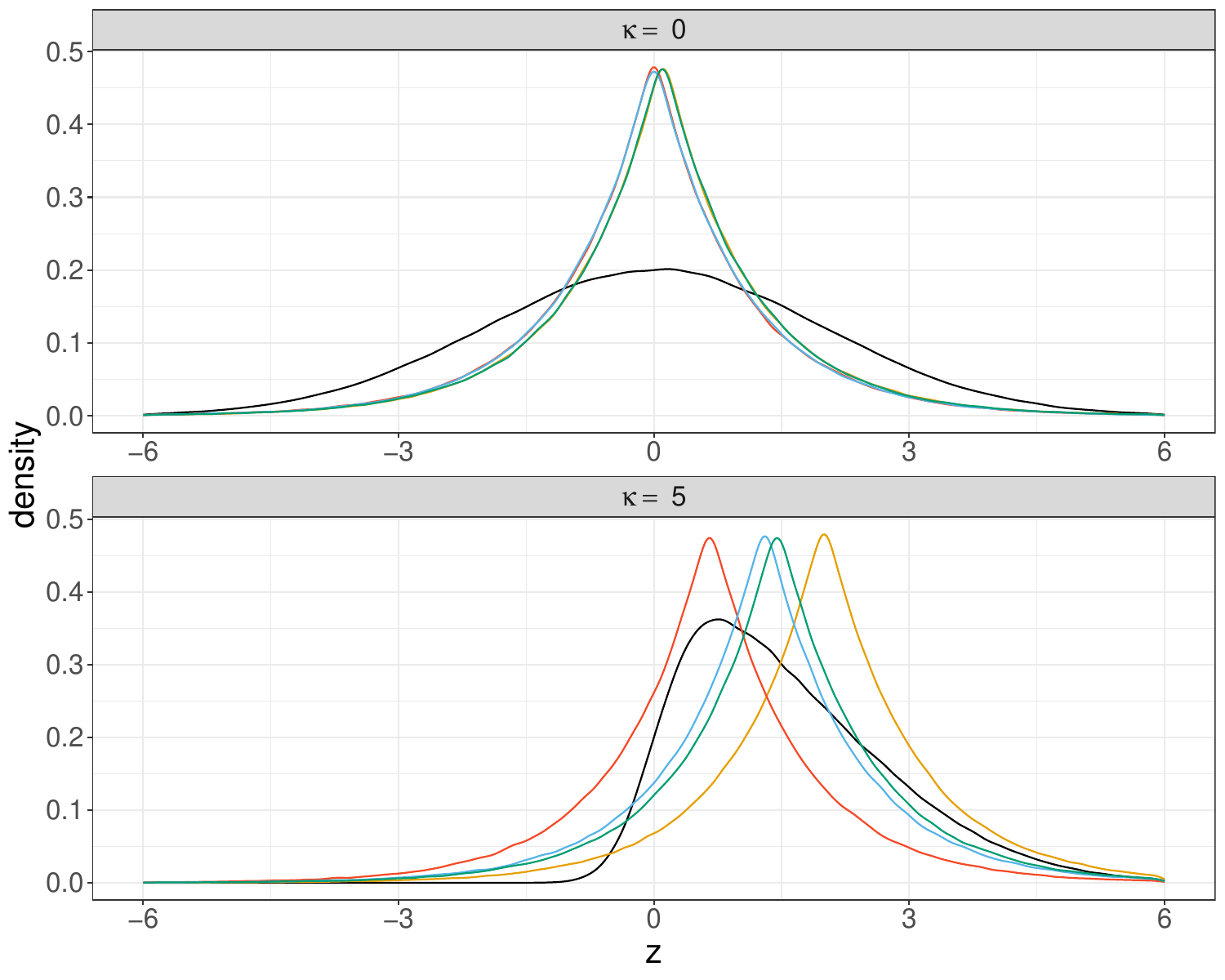}
    \includegraphics[width=3.5in]{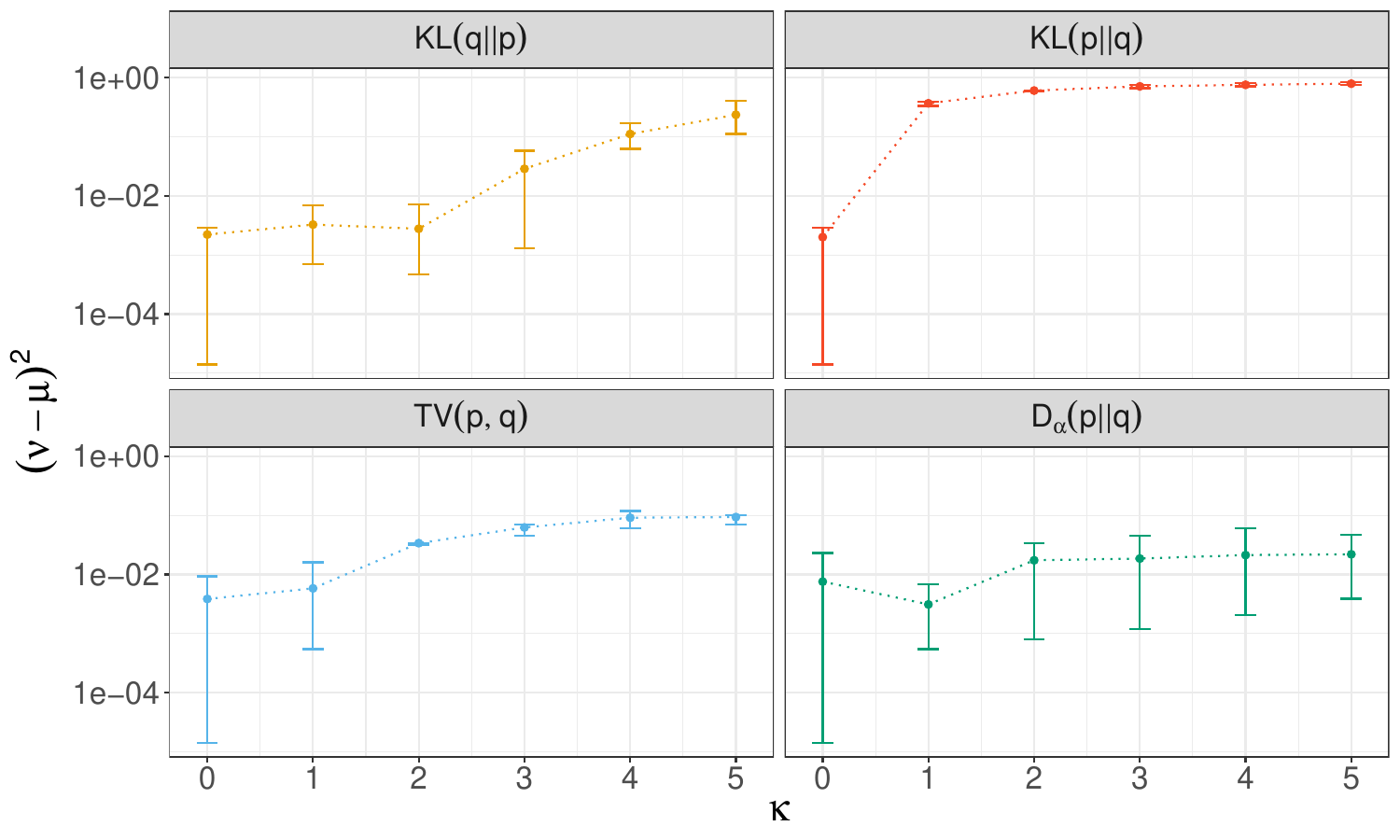}
    \caption{\textit{VI approximations to a skewed normal $p$ with a Laplace distribution. (Left) When $p$ has no skew $(\kappa\!=\!0)$, its mean is recovered by VI with all the divergences in \Cref{tab:divergences}; when $p$ is largely skewed $(\kappa\!=\!5)$, the results disagree. 
    (Right) The plot shows the squared error in the mean estimate averaged over 10 stochastic optimizations with the error bars spanning the 10$^\text{th}$ and 90$^\text{th}$ quantiles.}}
    \label{fig:skew-normal}
\end{figure*}

In this appendix, we provide an illustrative example of an \textit{asymmetric} target $p$. 
Let
\begin{eqnarray}
    p(z) = \text{skewed-}\mathcal{N}(0,2^2, \kappa), \ \ \
    q(z) = \text{Laplace}(\nu, 1),
\end{eqnarray}
where $\kappa\!\in\! [0, \infty)$ controls the skewness of $p$.
Here we chose $\mathcal Q$ to be the location-family of Laplace distributions,~so that the approximation remains misspecified, even when $\kappa\!=\! 0$ and the target is perfectly even symmetric.
Each divergence is evaluated via Monte Carlo and a minimizer is found using a grid-search.
For \mbox{$\kappa \!=\! 0$}, all divergences yield a solution that recovers the mean of $p$ (within some error in the stochastic grid search).
As~$\kappa$ increases, however, the error in the mean increases for all divergences, though not to the same extent; see \Cref{fig:skew-normal}.

\subsection{Existence of a stationary point for the variational scale matrix} \label[appendix]{app:radial-tail}

In this appendix, we provide a proof of \Cref{lemma:gamma-exists}, which shows that there exists a stationary point of $D_f(p||q)$ with respect to $\Gamma$, as defined in \cref{eq:Df_gl}.

We begin by characterizing the asymptotic behavior of the radial function $g_p$ in \Cref{remark:scalar_rep} when the elliptical density $p$ is unimodal. The next lemma shows that this behavior is strongly governed by the integrability of $p$.

\begin{bluebox}
\begin{lemma}[\textbf{Asymptotic decay of radial function $g_p$}] \label[lemma]{lemma:integrability}
    Let $p$ be an elliptically symmetric density over~$\mathbb{R}^d$ with radial function~$g_p$ as defined in \Cref{remark:scalar_rep}. If $g_p(r)$ is monotonically decreasing, then $g_p(r)\,r^d\rightarrow 0$ as $r\rightarrow\infty$.
\end{lemma}
\end{bluebox}

\begin{proof}
\noindent
We prove the lemma by contradiction. Suppose that $g(r)\,r^d\not\rightarrow 0$ as $r\!\rightarrow\!\infty$. Then there must exist some \mbox{$\epsilon\!>\!0$} for which we can produce arbitrarily large values of $r$ satisfying \mbox{$g_p(r)\,r^d\geq\epsilon$}. Let $\{r_1,r_2,\ldots,r_{K+1}\}$ denote an increasing sequence of such radii satisfying $r_{k}\! \geq\! 2 r_{k-1}$. Then we have the following chain of inequalities:
\begin{eqnarray}
\label{eq:radial_integral}
\int_{0}^\infty \!\!\!g_p(r)\, r^{d-1} dr
  & \geq & \sum_{k=1}^{K} \int_{r_k}^{r_{k+1}}\! g_p(r)\, r^{d-1}\, dr, \\
  & \geq & \sum_{k=1}^{K}\, \int_{r_k}^{r_{k+1}} g_p(r_{k+1})\,  r^{d-1}\, dr, \nonumber\\
  & \geq & \sum_{k=1}^{K}\, \int_{r_k}^{r_{k+1}} \left(\frac{\epsilon}{r_{k+1}^d}\right)\, r^{d-1}\, dr, \nonumber\\
  & = & \sum_{k=1}^{K}\,\frac{\epsilon}{r_{k+1}^d}\left(
   \frac{r_{k+1}^d - r_k^d}{d}\right), \nonumber\\
  & = & \frac{\epsilon}{d}\sum_{k=1}^{K}\,\left(1-\frac{r_{k}^d}{r_{k+1}^d}\right), \nonumber\\[0.5ex]
  & \geq & \frac{\epsilon K}{d}
  \left(1-\frac{1}{2^d}\right). 
  \label{eq:lower_bound}
\end{eqnarray}
On one hand, if $p$ is integrable over~$\mathbb{R}^d$, then the radial integral in \cref{eq:radial_integral} must be finite. On the other hand, the lower bound in \cref{eq:lower_bound} can be made arbitrarily large by increasing the number of $\epsilon$-violating radii in $\{r_1,r_2,\ldots,r_{K+1}\}$ and so we have a contradiction.
\end{proof}

We now prove \Cref{lemma:gamma-exists}.
In \Cref{lemma:gamma}, we showed that the stationary points of $D_f(p||q)$ are given by solutions of \cref{eq:Gamma-stationary}, which we restate here for convenience:
    \begin{equation*}
      0 = \int f' \left( \frac{g_p(||\Gamma z||)}{g_q(||z||)} |\Gamma| \right) \left[ g_p(||\Gamma z||) + g_p'(||\Gamma z||) \dfrac{z_i^2 \gamma_i^2}{||\Gamma z||} \right]  \text d z.
    \end{equation*}

We know from \Cref{lemma:gamma} that any stationary point of $D_f(p||q)$ must be of the form $\Gamma\! =\! \gamma I$. With this simplification, the above equation reduces to 
 \begin{equation}
 0 = \int f' \left( \frac{g_p(\gamma ||z||)}{g_q(||z||)} \gamma^d \right) \left[ g_p(\gamma||z||) + g_p'(\gamma||z||) \dfrac{z_i^2 \gamma}{||z||} \right] \text d z.
      \label{eq:intzi}
\end{equation}
Observe that by symmetry, the integral in \cref{eq:intzi} takes the same value for any choice of the coordinate $i$ (which appears only in the term involving $z_i^2$). We can exploit this symmetry by replacing the term $z_i^2$ in the integrand by $\|z\|^2/ d$. Making this substitution, we see that a stationary point of $D_f(p||q)$, of the form $\Gamma\!=\!\gamma I$, occurs at any solution $\gamma$ satisfying
  \begin{equation}
  0 = \int f' \left( \frac{g_p(\gamma ||z||)}{g_q(||z||)} \gamma^d  \right) \left[ g_p(\gamma||z||) + \frac{1}{d}\, g_p'(\gamma||z||)\, \gamma ||z|| \right] \text d z.
      \label{eq:phi}
\end{equation}
Our goal is to show that the integral in \cref{eq:phi} has a root for some positive value of $\gamma$, and we will do so by showing that the value of the integral has different signs in the opposing limits that $\gamma$ is very small and very large.
\begin{itemize}

  \item \fcolorbox{Cerulean!10}{Cerulean!10}{$\gamma \to 0$:} 
 In this limit we claim that the integral has the \textit{same} sign as  $\lim_{t\to0^+}f'(t)$. 
  To see this, note that as $\gamma\to 0$, the argument of $f'(\cdot)$ in the integrand of \cref{eq:phi} approaches zero from above, while the term inside square brackets approaches $g_p(0)\!>\!0$. Therefore the term in brackets \textit{preserves} the sign of the integrand as a whole.
  
  \item \fcolorbox{Cerulean!10}{Cerulean!10}{$\gamma \to +\infty$:}  In this limit we claim that the integral has the \textit{opposite} sign as  $\lim_{t\to0^+}f'(t)$. 
  %
  First, consider the term in the integrand involving $f'(\cdot)$. In the limit $\gamma\to +\infty$, we show that the argument of $f'(\cdot)$ once again vanishes from above. To see this, we rewrite the argument~as
  \begin{equation} \label{eq:argument-f-prime}
  \frac{g_p(\gamma\|z\|)}{g_q(\|z\|)}\gamma^d = \frac{g_p(\gamma\|z\|)\,(\gamma \|z\|)^d}{g_q(\|z\|)\|z\|^d}.
  \end{equation}
Next we draw on the fact that $p$ is integrable, from which it follows (as shown in \Cref{lemma:integrability}) that $\lim_{r\to\infty} g_p(r) r^d = 0$.
Setting $r=\gamma \|z\|$ in \cref{eq:argument-f-prime}, we see that the numerator vanishes as $\gamma\rightarrow\infty$, and thus the argument of $f'(\cdot)$ also vanishes.\footnote{More precisely, the argument of $f'(\cdot)$ vanishes everywhere in the limit $\gamma\to+\infty$ except at the origin of the integrand where $\|z\|=0$. But this isolated point of measure zero does not make a finite contribution to the integral.}

   Next, we consider the term in brackets.
  %
  %
  By assumption, $g_p(r) r^d$ is strictly decreasing in its tail. Computing its derivative, we find
  \begin{align} \label{eq:derivative-rg}
    \Big(g_p(r)\, r^d\Big)' =  d r^{d-1} \left[ g_p(r) + \frac{1}{d}\, g_p'(r) r \right].
  \end{align}
  Once again setting $r= \gamma ||z||$, we see that the bracketed term in \cref{eq:derivative-rg} corresponds \textit{exactly} to the bracketed term in the integrand of \cref{eq:phi}; hence
  the limit of the former as $r\!\rightarrow\!\infty$ matches
  the limit of the latter as $\gamma\rightarrow\infty$. 
  Moreover, by the strict monotonicity of the tails, the right side of \cref{eq:derivative-rg} in this limit is strictly negative, and then so is the bracketed term in \cref{eq:derivative-rg}. Therefore, in this limit the integral in \cref{eq:phi} has the \textit{opposite} sign of $\lim_{t\to0^+}f'(t)$.
\end{itemize}
In sum, we have shown that the integral in \cref{eq:phi} has opposite signs for values of $\gamma$ that are sufficiently small and sufficiently large.
By continuity, there must exist some $\gamma\!>\!0$ that yields a root of this equation and hence a stationary point of $D_f(p||q)$ at $\Gamma=\gamma I$.
This completes the proof.

\subsection{Proofs for Guarantees in the Presence of Partial Elliptical Symmetry} \label[appendix]{app:elliptical-sigma}

In this appendix, we provide supporting lemmas for the proof of \Cref{thm:elliptical-sigma}.
We begin by showing that if $q(\zs|\zb)$ matches the mean and scatter matrix of $p(\zs|\zb)$, then it must be a global minimizer of the average divergence,
\begin{equation} \label{eq:average-divergence}
        \overline{D}_{\sigma|\bsigma} = \int D_f \big [p(\zs|\zb)||q(\zs|\zb) \big ] h \big(p(\zb), q(\zb) \big) \text d \zb.
\end{equation}
(That such a minimizer can be constructed for any $q(\zb)$ is then shown in the proof of \Cref{thm:elliptical-sigma}.)
\begin{bluebox}
\begin{lemma}[{\bf Sufficient condition for the minimizer of an average divergence}] \label[lemma]{lemma:conditional-scatter}
Let $D_f(p||q)$ be an $f$-divergence which verifies the chain rule (\Cref{def:chain-rule}) with $f$ strictly convex,
let $\mathcal Q$ be an elliptical location-scale family,
and let $p$ be elliptically symmetric along $\sigma$ with a point of even symmetry $\nu_{\sigma|\bsigma}$ which is linear in $\zb$ and a constant scatter matrix $M_{\sigma|\bsigma}$.
In addition, suppose that
$p$ is strictly unimodal and its radial function along $\sigma$ $g_p(\cdot|\zb)$ is differentiable everywhere and strictly radially decreasing in its tails;
and any $q \in \mathcal Q$ is strictly unimodal and has a radial function $g_q$ which is everywhere continuous.
%

Then, if there exists $q \in \mathcal Q$ such that $q(\zs|\zb)$ matches the mean and scatter matrix of $p$, that is $\nu_{\sigma|\bsigma} = \mu_{\sigma|\bsigma}$ and $S_{\sigma|\bsigma} = \gamma^2 M_{\sigma|\bsigma}$ for all $\zb$, then $q(\zs|\zb)$ is a minimizer of \cref{eq:average-divergence}.
(Note: $\gamma$ is constant with respect to $\zb$.)
\end{lemma}
\end{bluebox}

\begin{proof}

$D_f\big [p(\zs|\zb)||q(\zs|\zb) \big]$ is minimized with respect to $\nu_{\sigma|\bsigma}$ for any $\zb$ by setting $\nu_{\sigma|\bsigma} = \mu_{\sigma|\bsigma}$.
In the proof of \Cref{thm:elliptical-sigma}, we show how to set $\nu_\sigma$ and $S_{\sigma \bsigma}$ to achieve this.

To obtain guarantees for the optimal conditional scatter matrix, we closely follow the steps in the proof of \Cref{thm:elliptical}.
Since $q(z)$ is elliptically symmetric, $q(\zs|\zb)$ is also elliptically symmetric with a constant scatter matrix $S_{\sigma|\bsigma}$.

First, we apply \Cref{lemma:congruence} to the conditional divergence and, using the fact that $\nu_{\sigma|\bsigma} = \mu_{\sigma|\bsigma}$, we obtain,
\begin{equation}
    \overline{D}_{\sigma|\bsigma} = \int \left [\int f \left ( \frac{g_p(||\Gamma \zs|| \mid \zb)}{g_q(||\zs|| \mid \zb)} |\Gamma| \right) g_q(||\zs|| \mid \zb) \text d \zs \right] h \big( p(\zb), q(\zb) \big) \text d \zb,
\end{equation}
where $\Gamma$ is a diagonal matrix and we write $\text{Diag}(\Gamma) = (\gamma_1, \gamma_2, \cdots, \gamma_d)$. 
Notice that, while $g_p(\cdot |\zb)$ and $g_q(\cdot | \zb)$ vary with $\zb$, $\Gamma$ is constant, since the conditional scatter matrices $S_{\sigma|\bsigma}$ and $M_{\sigma|\bsigma}$ are constant.

Next, we follow the steps in the proof of \Cref{lemma:gamma}.
We differentiate $\overline{D}_{\sigma|\bsigma}$ with respect to $\gamma_i$ and, adapting \cref{eq:Gamma-stationary}, we obtain,
\begin{equation}
    \int \left [  \int f' \left( \frac{g_p(||\Gamma \zs|| \mid \zb)}{g_q(||\zs|| \mid \zb)} |\Gamma| \right) \left[ g_p(||\Gamma \zs|| \mid \zb) + g_p'(||\Gamma \zs||\mid \zb) \dfrac{z_{\sigma, i}^2 \gamma_i^2}{||\Gamma \zs||} \right] \text d \zs \right] h \big( p(\zb), q(\zb) \big) \text d \zb = 0.
\end{equation}
This implies that, for any pair $(i, j)$,
\begin{align} \label{eq:index-equality-partial}
    \int \left [\int f' \left( \frac{g_p(||\Gamma \zs|| \mid \zb)}{g_q(||\zs|| \mid \zb)} |\Gamma| \right) g_p'(||\Gamma \zs|| \mid \zb) \dfrac{z_{\sigma, i}^2 \gamma_i^2 - z_{\sigma, j}^2 \gamma_j^2}{||\Gamma \zs||} \text d \zs \right]  h \big( p(\zb), q(\zb) \big) \text d \zb = 0.
\end{align}
If $\gamma_i < \gamma_j$, we have shown in the proof of \Cref{lemma:gamma} that the inner integral is strictly negative and this result must hold for any $\zb$.
Since, per the definition of the divergence chain rule, $ h \big( p(\zb), q(\zb) \big) > 0$, we have that the left side of \cref{eq:index-equality-partial} is strictly negative.
By contradiction, $\gamma_i = \gamma_j$ for any $(i, j)$ and any stationary point must have the form $\Gamma = \gamma I$.

To show the existence of a stationary point, we follow the steps of \Cref{lemma:gamma-exists}.
\end{proof}

Next, we rederive a somewhat standard result of probability, which states that the covariance between two variables can be rewritten as a covariance between a variable and an expectation value. 

\begin{bluebox}
\begin{lemma}[{\bf Covariance with conditional expectation}] \label[lemma]{lemma:covariance}
\begin{equation}
    \text{Cov}(\zb, \zs) = \text{Cov}(\zb, \mathbb E[\zs | \zb]).
\end{equation}
\end{lemma}
\end{bluebox}
\begin{proof}
We begin by applying Tower's law to the definition of the covariance,
\begin{eqnarray*}
    \text{Cov}(\zs, \zb) & = & \mathbb E \big [(\zs - \mathbb E [\zs])(\zb - \mathbb E [\zb]) \big]  \\
                         & = & \mathbb E \big [\mathbb E [(\zs - \mathbb E [\zs])(\zb - \mathbb E [\zb]) | \zb] \big].
\end{eqnarray*}
Notice that the second term in parentheses does not depend on $\zs$, and so it can be pulled out of the inner conditional expectation.
Then,
\begin{eqnarray*}
    \text{Cov}(\zs, \zb)  & = & \mathbb E \big [(\zb - \mathbb E [\zb]) \mathbb E[(\zs - \mathbb E [\zs]) | \zb] \big] \\
          & = & \mathbb E \big [(\zb - \mathbb E [\zb]) (\mathbb E[\zs | \zb] - \mathbb E [\zs]) \big] \\
          & = & \text{Cov}(\zb, \mathbb E[\zs | \zb]).
      \end{eqnarray*}
  \end{proof}

We now derive a lemma which characterizes the marginal and conditional mean and covariances of $p$, when $p$ is elliptically symmetric. 

\begin{bluebox}
\begin{lemma}[{\bf Constant conditional symmetry}] \label[lemma]{lemma:constant-symmetry}
    Suppose $p$ is elliptically symmetric along $\sigma$ with a constant point of even symmetry and a constant scatter matrix.
    Then $\mu_\sigma = \mu_{\sigma | \bar \sigma}$, $M_{\sigma \sigma} = M_{\sigma | \bar \sigma}$ and $\text{Cov}(\zs, \zb) = {\bf 0}$.
\end{lemma}
\end{bluebox}
\begin{proof}
    We show that $p(\zs)$ and $p(\zs | \zb)$ have matching elliptical symmetry.
    This follows from
  \begin{equation}
      p(\zs) = \int_{\bar \sigma} p(\zs | \zb) p(\zb) \text d \zb,
  \end{equation}
  and the fact that the elliptical symmetry of the integrand along $\sigma$ is constant with respect to $\zb$.
  Since the point of even symmetry is found at the mean, we therefore have $\mu_\sigma  = \mu_{\sigma | \bar \sigma}$.
  Similarly, from the matching elliptical symmetry, we have $M_{\sigma \sigma} = M_{\sigma | \bar \sigma}$.

  Combining \Cref{lemma:covariance} and the fact $\nu_\sigma = \nu_{\sigma | \bar \sigma}$, we obtain that
  \begin{equation}
      \text{Cov}(\zs, \zb) = \text{Cov}(\mathbb E[\zs], \zb) =  \text{Cov}(\mu_\sigma, \zb) = {\bf 0},
  \end{equation}
  since $\mu_\sigma$ does not depend on $\zb$.
\end{proof}

\subsection{Analysis of the Elliptical Funnel} \label[appendix]{app:funnel}

We now apply our theoretical results to the elliptical funnel and variations thereof.
We begin with \cref{eq:funnel}, restated here for convenience,
\begin{equation}
    \tau \sim \mathcal{N}(0, 1); \ \theta \sim \mathcal N(0, e^{2 \tau} M). \nonumber
\end{equation}
The joint distribution, $p(\tau, \theta) = p(\tau) p(\theta | \tau)$ is elliptically symmetric along $\theta$ with a constant point of even symmetry at 0 
and a constant scatter matrix.
It may seem surprising that the scatter matrix is constant, since the covariance of $\theta$ in $p(\theta | \tau)$ depends on $\tau$.
However, $\tau$ only controls the scale of the covariances and does not alter the elliptical symmetry itself.
In particular, the conditional scatter matrix is $M$ for any value of $\tau$.
One can further show that $\tau$ and $\theta$ are uncorrelated, and that the marginal scatter matrix of $\theta$ is still $M$, per \Cref{lemma:covariance}.
Moreover, we have from \Cref{cor:elliptical-sigma} that VI recovers the \textit{marginal} mean and correlation matrix of $\theta$.

Consider now the elliptical funnel with a varying mean $\mu \in \mathbb R$,
\begin{equation} \label{eq:funnel-mean}
    \mu \sim \mathcal N(0, 1); \ \tau \sim \mathcal N(0, 1); \ \theta \sim \mathcal N(\mu, e^{2 \tau} M).
\end{equation}
$p(\mu, \tau, \theta)$ remains even symmetric along $\theta$, however the point of even symmetry is now given by $\mu$ and is no longer constant.
Since the point of even symmetry depends linearly on $\mu$, we obtain from \Cref{thm:elliptical-sigma} that VI recovers the \textit{conditional} mean and correlation for $\theta$.
Next, we may recognize that $p(\mu, \theta | \tau)$ is Gaussian and therefore $p$ is elliptically symmetric along $(\mu, \theta)$ with a constant point of even symmetry at 0.
However, the scatter matrix of $p(\mu, \theta | \tau)$ is \underline{not} constant.
This can be seen by examining the correlation between $\mu$ and $\theta$, which depends on $\tau$ (and goes to 1 as $\tau \to -\infty$).
We can therefore not apply our theoretical results and do not have guarantees on how well VI estimates the mean and correlation matrix of $(\mu, \theta)$.

A similar reasoning applies when considering a nested elliptical funnel,
\begin{equation}
    m \sim \mathcal N(0, 1); \ s \sim \mathcal N(0, 1); \ \mu \sim \mathcal N(m, e^{2s}); \ \tau \sim \mathcal N(0, 1); \  \theta \sim \mathcal N(\mu, e^{2 \tau} M),
\end{equation}
which characterizes the hierarchical prior that may arise in a nested multilevel model.
Noting that $p(\theta | \mu, \tau, m, s) = p(\theta | \mu, \tau)$, we have that $p(\theta, \mu, \tau, m, s)$ is partially elliptically symmetric along $\theta$ with a point of even symmetry that is linear in $\mu$ and a constant scatter matrix $M$.
Hence, \Cref{thm:elliptical-sigma} applies and $q(\theta | \mu, \tau, m, s)$ recovers the mean and scatter matrix of $p(\theta | \mu, \tau, m, s)$.
But this result does not extend to $\mu$, because the scatter matrix of $p(\mu|\theta, \tau, m, s)$ depends on $\tau$ and $s$, and similarly it does not extend to $m$ the scatter matrix of $p(m|\mu, \theta, \tau, s)$ depends on $s$.

Finally, we consider the more general funnel, 
where $M$ is allowed to vary with a hyperparameter $\rho \in \mathbb R$, 
%
\begin{equation}
    \mu \sim \mathcal N(0, 1); \ \tau \sim \mathcal N(0, 1); \ \rho \sim p(\rho); \ \theta \sim \mathcal N(\mu, e^{2 \tau} M(\rho)).
\end{equation}
In this setting, the scatter matrix is not constant and so we cannot apply our theoretical results.


\section{Experimental Details} \label[appendix]{app:details}

In this appendix, we provide additional details for the numerical experiments in \Cref{sec:experiments}.

The code to reproduce all experimental results and figures in the paper is available at \url{https://github.com/charlesm93/VI_elliptical_symmetry}.
We use {\sc R} as a scripting language and {\sc Stan} \citep{Carpenter:2017, Stan:2025} as a probabilistic programming language to specify models and run VI and MCMC.
Our work with {\sc Stan} is greatly facilitated by the packages {\sc BridgeStan} \citep{Roualdes:2023}.
All experiments are run on CPU using a 2.8 GHz Quad Core Intel Core i7 processor. 

\subsection{Targets} \label[appendix]{app:targets}

Here, we provide details on the targets in \Cref{tab:targets}.
We specify the coordinates $\sigma$ along which the target is even symmetric for the synthetic targets (\texttt{student-t}, \texttt{funnel}, and \texttt{crescent}) and \textit{a priori} even symmetric for the Bayesian models (\texttt{schools}, \texttt{disease}, \texttt{SKIM}).
Symmetry in the prior can manifest as \textit{approximate} symmetry in the posterior.

\fcolorbox{Cerulean!10}{Cerulean!10}{\texttt{student-t} ($d=2$).} A multivariate Student-t distribution with correlation 0.5.
The target is elliptically symmetric and $\sigma = (z_1, z_2)$.

\fcolorbox{Cerulean!10}{Cerulean!10}{\texttt{funnel} ($d = 4$).} The elliptical funnel with varying mean, specified by \cref{eq:funnel-mean}.
The correlation matrix $C$ in $p(\theta | \mu, \tau)$ has off-diagonal element $C_{12} = 0.5$.
Here $\sigma = (\mu, \theta)$.

\fcolorbox{Cerulean!10}{Cerulean!10}{\texttt{crescent} ($d = 3$).} The elliptical Rosenbrock distribution is comprised of a two-dimensional Gaussian and a third coordinate that depends quadratically on the first two components.
When plotted, the joint density between the third component and any of the first two components has the shape of a crescent.
In details, for $x \in \mathbb R^2$ and $y \in \mathbb R$,
\begin{equation}
    x \sim \mathcal N(0, \Sigma); \ \ y \sim \mathcal N(a (||x||_{\Sigma^{-1}} - b), c^2).
\end{equation}
In this experiment, we set
\begin{equation}
\Sigma = 10^2
\begin{pmatrix}
    1 & 0.5 \\ 0.5 & 1
\end{pmatrix},
\ a = 0.03, \ b = 100, \ c = 0.02.
\end{equation}
This distribution is even and elliptically symmetric along its first two coordinates, $\sigma = x$.
This is an extension of the two-dimensional Rosenbrock distribution by \citet{Roberts:1997}.
The two-dimensional Rosenbrock distribution is even-symmetric along its first coordinate.
We add an additional dimension in order to obtain a non-trivial partial elliptical symmetry.

\fcolorbox{Cerulean!10}{Cerulean!10}{\texttt{schools} $(d = 10)$.} A Bayesian hierarchical model of the effects of a preparation program for a standardized test across $N = 8$ schools \citep{Rubin:1981, Gelman:2013}.
We observe $y_i$, the average change in test scores, and $\eta_i$, the empirical standard deviation across students, for each school.
The model is then
\begin{equation} \label{eq:schools}
    \mu \sim \mathcal N(5, 3^2); \ \tau \sim \mathcal N^+(0, 5^2); \ \theta_i \sim \mathcal N(\mu, \tau^2); \ y_i \sim \mathcal N(\theta_i, \eta^2_i).
\end{equation}
The prior $p(\mu, \tau, \theta)$ is even symmetric along $\sigma = (\mu, \theta)$.
We can implement this model either using the standard (centered) parameterization (\cref{eq:schools}), the non-centered parameterization described in \Cref{sec:sym}, or by marginalizing out $\theta$---here we exploit the fact the prior $p(\theta | \mu, \tau)$ and likelihood $p(y_i | \theta_i)$ are Gaussians, and so,
\begin{equation}
    p(y_i | \mu, \tau) = \mathcal N(0, \sigma^2_i + \tau^2); \ p(\theta_i | y_i, \mu, \tau) = \mathcal N \left ( \frac{y_i / \eta_i^2 + \mu / \tau^2}{1/\eta_i^2 + 1/\tau^2}, \frac{1}{1/\eta_i^2 + 1/\tau^2} \right).
\end{equation}

\fcolorbox{Cerulean!10}{Cerulean!10}{\texttt{disease} $(d = 102)$.} A model for the mortality counts across counties in Finland, due to alcoholism \citep{Vanhatalo:2019}.
For each county, we observe the mortality count, $y_i$, the standardized expected number of deaths, $y_{e,i}$, and the two-dimensional location of the county, $x_i$.
The original model considers 911 counties, however we consider a random subset of 100 counties to reduce the computational cost of the experiment.
The model uses a Gaussian process prior with a squared exponential kernel.
Specifically, the prior covariance matrix $K$ is defined by,
\begin{equation}
    K_{ij} = \alpha^2 \exp \left (-\frac{||x_i - x_j||^2}{\rho^2} \right),
\end{equation}
and the full model is,
\begin{align}
    \rho \sim \text{inv-Gamma}(2.42, 14.81); \ \alpha \sim \text{inv-Gamma}(10, 10); \nonumber \\ \theta \sim \mathcal N(0, K(\alpha, \rho, x)); \ y_i \sim \text{Poisson}(y_{e, i} \exp(\theta_i)). 
\end{align}
The prior is even symmetric along $\sigma = \theta$.
As before, this model can be implemented using a centered or non-centered parameterization.
Exact marginalization is not possible, but can be achieved using an integrated Laplace approximation. 

\fcolorbox{Cerulean!10}{Cerulean!10}{\texttt{SKIM} $(d = 305)$.} A sparse kernel interaction model (SKIM) \citep{Agrawal:2019}.
This model is a regularized regression model that accounts for interaction effects between covariates.
Covariates are probabilistically selected using a horseshoe prior \citep{Piironen:2017}.
As in \citet{Margossian:2020}, we apply the model to a genetic microarray classification data set on prostate cancer.
We observe $N=102$ patients with $p=200$ pre-selected genetic covariates (out of a total of 5966 covariates) and denote $X \in \mathbb R^{N \times p}$ the design matrix.
We observe for each patient $y_i$, a binary variable that indicates whether the patient has cancer.

To specify the Bayesian model, we first set the following hyperparameters:
\begin{eqnarray}
    p_0 = 5; \ 
    s_\mathrm{global} = \frac{p_0}{\sqrt{N}(p - p_0)}; \ 
    \nu_\mathrm{local} = 1; \
    \nu_\mathrm{global} = 1; \
    s_\mathrm{slab} = 2 \
    s_\mathrm{df} = 100 \
    c_0 = 5.
\end{eqnarray}
Then, a standard parameterization of the model is,
\begin{eqnarray}
  \lambda_i \sim \mathrm{Student}_t(\nu_\mathrm{local}, 0, 1); \
     \tau \sim  \mathrm{Student}_t(\nu_\mathrm{global}, 0, s_\mathrm{global});  \
     c_\mathrm{aux} \sim \mathrm{invGamma}(s_\mathrm{df} / 2, s_\mathrm{df} / 2); \nonumber \\
  \chi \sim \text{InverseGamma}(s_\mathrm{df} / 2, s_\mathrm{df} / 2); \
  c = \sqrt{c_\text{aux}} s_\text{slab}; \
  \tilde \lambda_i^2 =  \frac{c^2 \lambda_i^2}{c^2 + \tau^2 \lambda_i^2}; \ 
 \eta_2 = \tau^2 \chi / c^2 \nonumber \\
 \beta_0 \sim \mathcal N(0, c_0^2); \
\beta_{i} \sim \mathcal N(0, \tau^2 \tilde{\lambda}_i^2);  \
\beta_{ij} \sim \mathcal N(0, \eta_2^2 \tilde{\lambda}_i^2 \tilde{\lambda}_j^2); \
y \sim \text{Bernoulli}(\text{logit}^{-1}(\beta_0 + X \beta)).
\end{eqnarray}
\textcolor{teal}{IR: the (62) looks snug.}

Following \citet{Agrawal:2019}, we marginalize out $\beta_i$ and $\beta_{ij}$ using a Gaussian process reparameterization.
To define the Gaussian process' covariance matrix $K$, we first introduce the matrices:
\begin{eqnarray}
    K_1 & = & X \ \mathrm{diag}(\tilde{\lambda}^2) \ X^T \nonumber \\
    K_2 & = & [X \circ X] \ \mathrm{diag}(\tilde{\lambda}^2) \ [X \circ X]^T,
  \end{eqnarray} 
   where ``$\circ$'' denotes the element-wise Hadamard product.
   Finally,
   \begin{eqnarray}
    K & = & \frac{1}{2} \eta_2^2 (K_1 + 1) \circ (K_1 + 1) - \frac{1}{2} \eta_2^2 K_2
    - (\tau^2 - \eta_2^2) K_1 + c_0^2  - \frac{1}{2} \eta_2^2.
\end{eqnarray}
Then, the Gaussian process prior and the likelihood are
\begin{equation}
    f \sim \mathcal N(0, K); \ y \sim \text{Bernoulli}(\text{logit}^{-1}(f)).
\end{equation}
Once again, this model admits three implementations: a centered parameterization, a non-centered parameterization, and an implementation where $f$ is approximately marginalized out with an integrated Laplace approximation.

\subsection{VI algorithm} \label[appendix]{app:algorithm}

For the experiment in \Cref{sec:experiments}, we specify targets in {\sc Stan} and fit them with ADVI \citep{Kucukelbir:2017}.
ADVI employs a Gaussian approximation over the unconstrained scale.
Constrained variables are automatically transformed to an unconstrained scale by {\sc Stan}.
For example, a variable $z \in (0, \infty)$ is replaced by $\log z \in \mathbb R$. 
In our experiments, we report estimates of the mean on the unconstrained scale, since our theoretical analysis applies to variables defined over $\mathbb R$.
This choice allows us to test how predictive/illustrative our theory is. While it is admittedly of more practical interest to estimate the mean over the original scale, 
our theoretical results do not apply there.

ADVI minimizes $\text{KL}(q||p)$ via stochastic optimization.
We warm-start VI using a factorized (mean-field) approximation, then switch to a Gaussian with a full covariance matrix.
We use a large batch size ($B\!\ge\! 50$) to better estimate the ELBO and its gradient, and thereby improve our chances of finding an optimal solution.
All other tuning parameters are set to the default options in {\sc Stan}.

\subsection{Additional results on correlation and covariance} \label[appendix]{app:correlation}

\begin{table*}[]
    \centering
    \small
    \begin{tabular}{l r r r r}
    & & & (nc) & (marg) \\[-0.05in]
    \textbf{Call name} & $d$ &  $b_\text{90}$ & $b_\text{90}$ & $b_\text{90}$ \\
    \rowcolor{gray!20} \texttt{student} & 2 & 0 & & \\
    \texttt{crescent} & 3 & 29.36 & & \\
    \rowcolor{gray!20} \texttt{funnel} & 4 & 112.53 & & \\
    \texttt{schools} & 10 & 92.63 & 2.73 & 1.07 \\
    \rowcolor{gray!20} \texttt{disease} & 102 & 80,091 & 71.75 & 0.61 \\
    \texttt{SKIM} & 305 & 64,958 & 104.38 & 844.63
    \end{tabular}
    \caption{\textit{Target distributions for experiments and estimated elliptical asymmetry $b_\text{90}$; see \cref{eq:symmetry}.
    The hierarchical models (\texttt{schools}, \texttt{disease}, \texttt{SKIM}) were implemented using (i) a standard centered parameterization, (ii) a non-centered parameterization with less asymmetry, (iii) an approximate marginalization of the latent variables, yielding a collapsed (and even less asymmetric) posterior; see the discussion in \Cref{sec:sym}.}}
    \label{tab:targets-ellipse}
\end{table*}

As in \Cref{sec:experiments}, we order the targets by their asymmetry, this time evaluating their \textit{elliptical} asymmetry.
Given a sample $z$ from $p(z)$, we compute a random ``reflection'' as follows.
Let $\hat \mu$ be the benchmark estimate of the mean of $p$ and let $\hat \Sigma$ be the benchmark estimate of the covariance matrix.
Furthermore, let $R$ be a random orthonormal matrix drawn from a Haar measure.
Then, we take,
\begin{equation}
    z' = \hat \Sigma^\frac{1}{2} R \hat \Sigma^{-\frac{1}{2}} (z -\hat \mu) + \hat \mu.
\end{equation}
One can check that $(z' - \hat \mu)^T \hat \Sigma^{-1}(z'-\hat \mu) = (z-\hat \mu)^T \hat \Sigma^{-1}(z-\hat \mu)$, meaning that $z$ and $z'$ lie on the same elliptical contour.
Hence, if $p$ is elliptically symmetric (about $\hat \mu$, with scatter matrix proportional to $\hat \Sigma$), then $p(z) = p(z')$.
The measure of asymmetry is then $b(z) = |\log \pi(z, x) - \log \pi(z', x)|$.

We evaluate $b(z)$ for 20,000 reference samples (either obtained exactly or via MCMC) and report the $90^\text{th}$ quantile.
For \texttt{SKIM}, we find the cost of evaluating $z'$ and $b(z)$ steep and only use 2,000 samples.
As before, values of $b(z)$ which evaluate to \texttt{NAN} are simply ignored
(this occurs in \texttt{SKIM} in $\sim$7.5\% of cases with an integrated Laplace and in $<1\%$ of cases with a centered parameterization).
\Cref{tab:targets-ellipse} reports the elliptical asymmetry of each target.
The relative elliptical asymmetry is comparable to what we obtain for even-asymmetry (\Cref{tab:targets}).

\begin{figure}[h]
    \centering
    \includegraphics[width=0.75\linewidth]{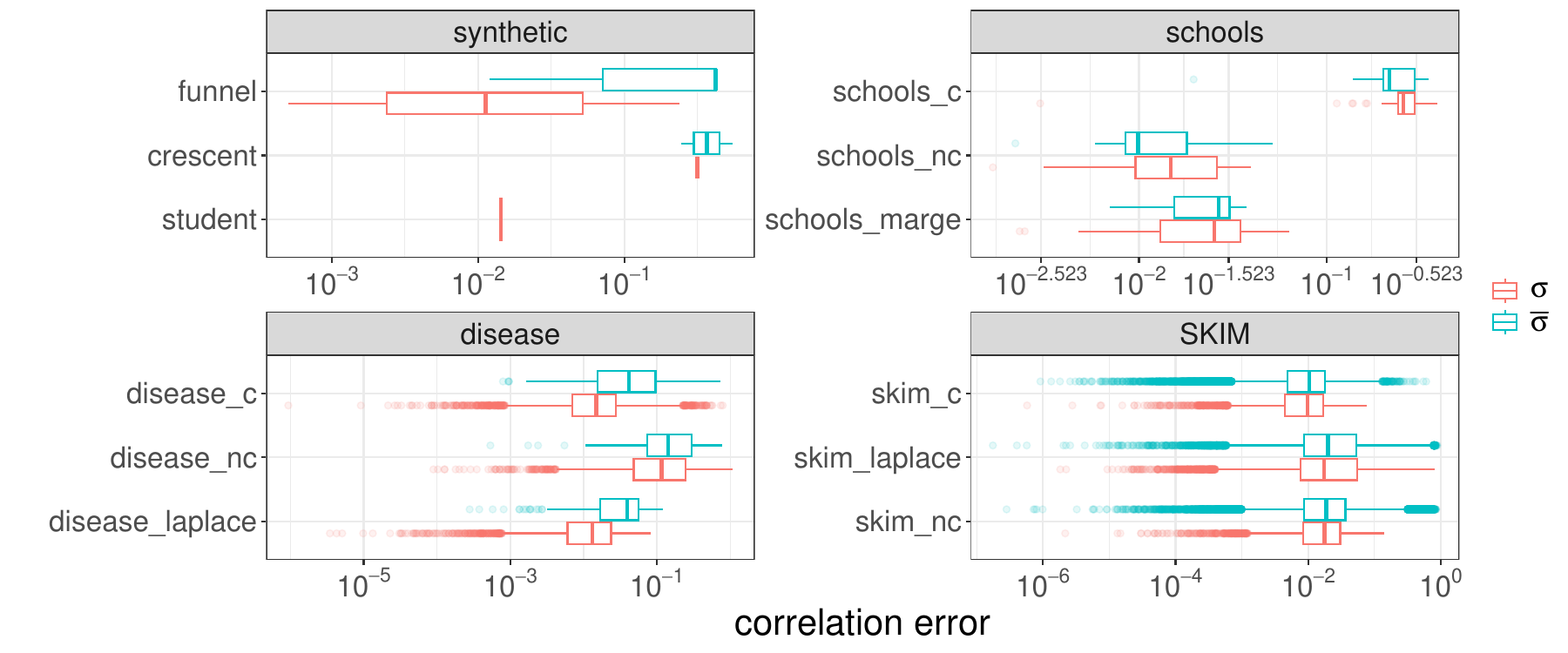}
    \caption{\textit{Absolute errors in VI estimates of the correlation.
    We split the targets into four groups: synthetic targets and implementations of \texttt{schools}, \texttt{disease}, and \texttt{SKIM}.
    Within each panel, the models are ordered bottom to top from most elliptical symmetric to least elliptical symmetric.
    }}
    \label{fig:corr}
\end{figure}

\Cref{fig:corr} plots the distribution of errors in estimates of the correlations across the models.
The models are ordered according to their elliptical asymmetry.
We also identify coordinates of partial elliptical symmetry.
Overall, we do not find a clear trend in how the correlation error changes as a function of elliptical asymmetry.
We suspect this is because many of the exact correlations are close to 0, especially in \texttt{disease} and \texttt{SKIM}.
In the synthetic examples, we obtain slightly better correlation error along coordinates of partial elliptical symmetry.
In \texttt{schools}, the non-centered and marginalized parameterization, with a reduced elliptical asymmetry, produce better estimates than the centered parameterization.



\begin{figure}[h]
    \centering
    \includegraphics[width=0.75\linewidth]{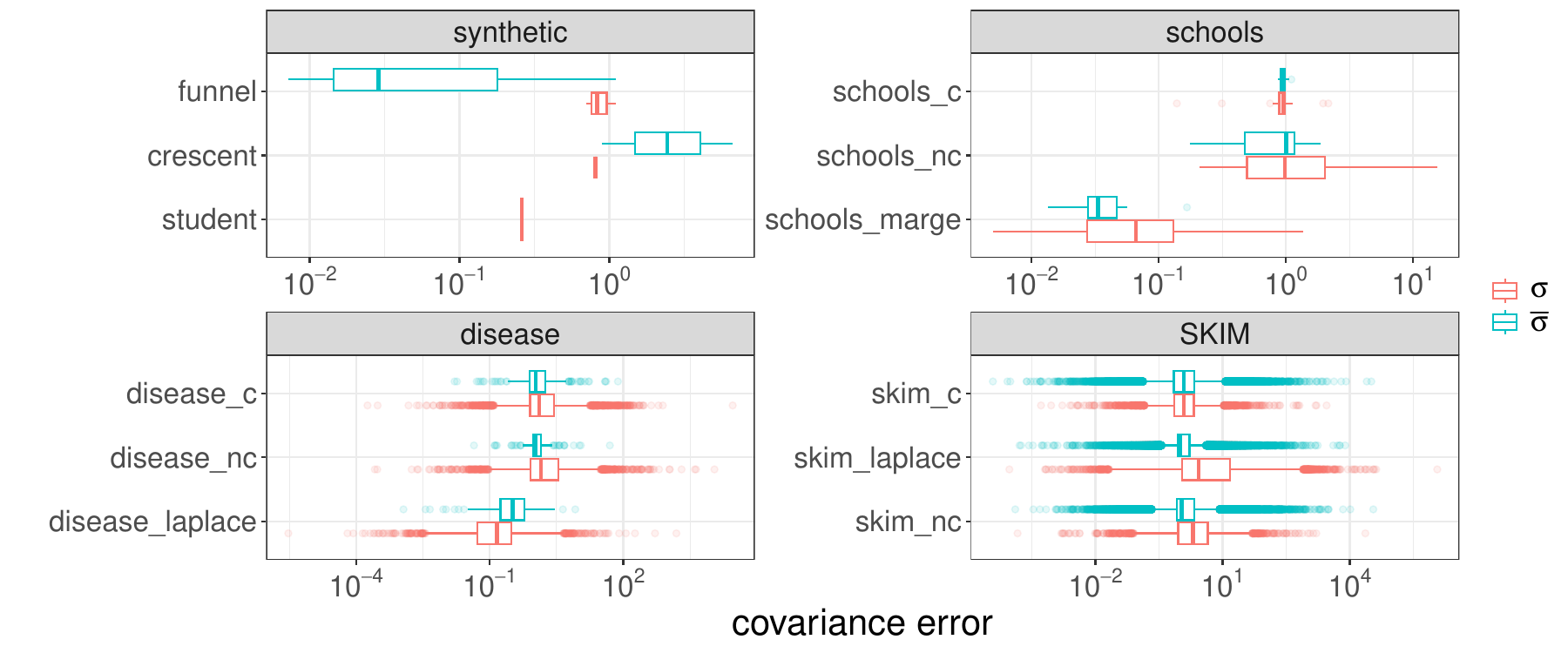}
    \caption{\textit{Fractional error in VI estimates of the covariance.
    We split the targets into four groups: synthetic targets and implementations of \texttt{schools}, \texttt{disease}, and \texttt{SKIM}.
    Within each panel, the models are ordered bottom to top from most elliptical symmetric to least elliptical symmetric.
    }}
    \label{fig:cov}
\end{figure}

Finally, we examine the error in the covariance estimate.
Our theory provides guarantees for the recovery of the correlations but not the covariances, which are 
only recovered up to a multiplicative constant.
\Cref{fig:cov} plots the distribution of fractional errors in the covariances, $\text{Err}_{ij} = |\Psi_{ij} - \hat \Sigma_{ij}| / |\hat \Sigma_{ij}|$, where $\Psi$ is the variational covariance matrix and $\hat \Sigma$ the benchmark covariance.
This figure contrasts with \Cref{fig:corr} and we highlight two examples where the correlations are accurately recovered but the covariances are not:
(i) for \texttt{student}, the mean correlation error is $\sim$0.014, while the mean fractional covariance error is $\sim$0.25;
and (ii) for \texttt{schools} with a non-centered parameterization, the mean correlation error is $\sim$0.015, while the mean fractional covariance error is $\sim$1.47.

\end{document}